\documentclass{article}

\PassOptionsToPackage{numbers, compress}{natbib}
\usepackage[preprint]{neurips_2026}

\usepackage[utf8]{inputenc} % allow utf-8 input
\usepackage[T1]{fontenc}    % use 8-bit T1 fonts
\usepackage{hyperref}       % hyperlinks
\usepackage{url}            % simple URL typesetting
\usepackage{booktabs}       % professional-quality tables
\usepackage{amsfonts}       % blackboard math symbols
\usepackage{nicefrac}       % compact symbols for 1/2, etc.
\usepackage{microtype}      % microtypography
\usepackage{xcolor}         % colors
\usepackage{amsmath}
\usepackage{array}
\usepackage[normalem]{ulem}
\usepackage[table]{xcolor}
\usepackage{subcaption}
\usepackage{graphicx}
\usepackage{multirow}
\usepackage{enumitem}

\title{Backdoor Mitigation in Object Detection via Adversarial Fine-Tuning}

\usepackage{authblk}

\author[1]{\textbf{Kealan Dunnett}}
\author[2]{\textbf{Reza Arablouei}}
\author[1]{\textbf{Dimity Miller}}
\author[2]{\textbf{Volkan Dedeoglu}}
\author[1]{\textbf{Raja Jurdak}}

\affil[1]{Queensland University of Technology, Brisbane Australia}
\affil[2]{CSIRO, Pullenvale QLD, Australia}

\begin{document}

\maketitle

\begin{abstract}
Backdoor attacks can implant malicious behaviours into deep models while preserving performance on clean data, posing a serious threat to safety-critical vision systems. Although backdoor mitigation has been studied extensively for image classification, defenses for object detection remain comparatively underdeveloped. Adversarial fine-tuning is a common backdoor mitigation approach in classification, but adapting it to detection is nontrivial as classification-oriented adversarial generation does not match the detection attack space, where attacks may cause object misclassification or disappearance, and standard detection losses can dilute the repair signal across many predictions. We address these challenges through a detection-aware adversarial fine-tuning framework for mitigating object-detection backdoors when the defender has access only to a compromised detector and a small clean dataset, without knowing the attack objective.
For adversarial generation that does not require knowledge of the attack objective, we introduce soft-branch minimisation, which uses a soft gate to combine objectives aligned with misclassification and disappearance attacks, together with a detection-aware classification-loss maximisation. For targeted repair, we introduce a dual-objective fine-tuning loss applied to target-matched predictions, concentrating the defensive update on predictions most relevant to the backdoor behaviour. Experiments across CNN- and Transformer-based detectors show that our approach more effectively reduces attack success while preserving true detections, compared with classification-oriented baselines, and maintains competitive clean detection performance.
\end{abstract}

\section{Introduction}\label{sec:intro}

The rapid deployment of deep learning in vision systems has heightened concerns about security vulnerabilities that can directly compromise downstream decisions in the physical world~\cite{liu2020privacy}. Among the most concerning threats are backdoor attacks. In a backdoor attack, an adversary tampers with the training pipeline (e.g., by poisoning a small subset of the training samples) so that the trained model behaves normally on clean inputs but exhibits attacker-chosen behavior whenever a specific \emph{trigger} pattern is present (e.g., a small colored patch)~\cite{gu2019badnets}. Although backdoors have been studied extensively in image classification, their behavior and mitigation in object detection remain much less understood, despite the central role of detectors in safety-critical applications such as autonomous driving and surveillance.

Backdoors in object detection differ fundamentally from those in classification. Rather than producing a single class, modern detectors produce a set of predictions, namely bounding boxes with associated class scores, which expands both the attack surface and the range of possible attack objectives. Among the threat models studied in prior work~\cite{chan2022baddet,luo2023untargeted,dunnett2026baddet+}, object targeting attacks (i.e., attacks that target the predictions of specific objects in an image) are particularly prominent. In region misclassification attacks (RMAs), trigger-bearing objects are misclassified into an attacker-chosen target class. In object disappearance attacks (ODAs), the predictions for attacked objects are suppressed~\cite{chan2022baddet}.

%Recent work (e.g., BadDet+ \cite{dunnett2026baddet+}) has shown that both RMA and ODA objectives can be induced within a unified framework while largely preserving clean performance, highlighting a practical and severe vulnerability. 
However, defenses for backdoor attacks in object detection have not kept pace. Existing countermeasures are dominated by two directions: (i) inference-time trigger detection~\cite{zhang2025test} and (ii) trigger synthesis or pattern-recovery approaches~\cite{shen2023django,cheng2024odscan}. By contrast, direct backdoor mitigation (i.e., repairing a compromised detector using only limited clean data) remains largely unexplored. While \cite{zhang2024towards} proposes a mitigation method, their approach is restricted to older two-stage detectors (e.g., Faster R-CNN), and assumes access to clean and backdoor data. As a result, methods that propose a more general-purpose design applicable to a wider range of detector classes without assuming access to backdoor data are required moving forward.

%While backdoor mitigation using only clean data is relatively mature in image classification \cite{wu2022backdoorbench,dunnett2024countering}, it remains significantly underexplored for object detection. 
Among classification-oriented approaches, adversarial fine-tuning has emerged as a prominent mitigation paradigm, using adversarial examples as surrogate triggers to identify and suppress the vulnerable pathways exploited by a backdoor~\cite{wei2023sau,zhu2023enhancing,zeng2021adversarial}. However, adapting this paradigm to object detection is not straightforward. Object detectors produce multiple coupled predictions with intertwined localization and classification components, and the relevant attack behaviours are heterogeneous. RMA requires a non-target class score to cross the detection threshold, while ODA requires all foreground class scores to fall below it. 

A second difficulty arises during fine-tuning. Detection losses are aggregated across many candidate predictions, most of which correspond to background regions or non-target objects. When fine-tuning on an adversarially perturbed image, the gradient associated with the targeted object can therefore be diluted by contributions from unrelated predictions, limiting the effectiveness of the standard detection loss as an object-level repair signal. These observations suggest that adversarial fine-tuning for detection requires two detection-specific components: (i) adversarial generation aligned with the score-threshold structure of both RMA and ODA without requiring the defender to know which the adversary is using, and (ii) a concentrated repair objective focused on predictions matched to the targeted object.

Motivated by this, we develop a detection-aware adversarial fine-tuning approach for backdoor mitigation in object detection. Our approach addresses two structural mismatches between classification and detection settings: (i) adversarial generation must handle an unknown attack objective, and (ii) the repair signal must concentrate on the targeted object. To address the first, we introduce soft-branch minimisation (SBM), which softly combines RMA- and ODA-aligned objectives when the attack objective is unknown to the defender. We contrast SBM with classification-loss maximisation (CLM), a detection-aware adaptation of the standard maximisation paradigm. To address the second, we propose a dual-objective defence loss applied to target-matched predictions. This loss complements the standard detection loss by directly restoring the ground-truth class score while suppressing competing class scores at the object level. We evaluate the approach across CNN- and Transformer-based detectors (FCOS, YOLOv5, DINO) on Pascal VOC, MTSD, and COCO, and against both RMA and ODA. The results show that our approach improves the trade-off between attack success rate (ASR) and true detection rate (TDR) relative to classification-oriented baselines while maintaining a favourable clean detection performance.

% In this work, we propose a general-purpose framework for backdoor mitigation that applies adversarial fine-tuning to a broad class of detection models. Our contributions are threefold:

% \begin{itemize}[leftmargin=6mm]
%     \item \emph{Algorithmic reformulation:} We revisit standard maximization-based adversarial example generation for detection, propose classification-loss maximisation (CLS) and introduce an alternative soft branch minimisation (SBM) method, tailoring adversarial generation to the structural complexities of detection models
%     \item \emph{Detection-aware repair objective:} We design a training objective that exploits the multi-prediction structure of detectors to produce informative repair gradients, allowing for precise isolation and removal of the backdoor behaviour.
%     \item \emph{Empirical validation:} We demonstrate the effectiveness of the proposed approach across diverse CNN- and Transformer-based detectors, achieving substantial mitigation while largely preserving clean detection performance (mAP).
% \end{itemize}

\section{Background and Related Work}\label{sec:related-work}

\textbf{Backdoor Attacks in Object Detection:}
BadDet~\cite{chan2022baddet} introduced backdoor attacks for object detection and formalized threat models such as RMA and ODA. Subsequent work expanded these attack behaviors and improved their practicality, for example, by extending disappearance to the untargeted setting~\cite{luo2023untargeted}, inducing disappearance without explicit annotation editing~\cite{cheng2023attacking}, and evaluating more realistic scene-level manipulations and real-world benchmarks~\cite{doan2024credibility}. More recently, BadDet+~\cite{dunnett2026baddet+} revisited this line of work and showed that existing data-poisoning attacks are substantially less effective than originally reported: BadDet remains a viable RMA attack but frequently produces duplicate detections in which the original class persists alongside the target, while existing ODA attacks~\cite{luo2023untargeted, cheng2023attacking, doan2024credibility} fail to reliably suppress objects even at extreme poisoning ratios. To address this, BadDet+ proposes a unified training-time loss manipulation that reliably achieves both RMA and ODA across detectors and datasets.

\textbf{Backdoor Defenses for Object Detection:}
Compared with image classification, defenses tailored to backdoors in object detection remain limited. Existing efforts fall into two main directions: (i) \emph{test-time detection} of triggered inputs or objects, e.g., by checking prediction consistency under input transformations~\cite{zhang2025test}; and (ii) \emph{trigger inversion or scanning} methods that synthesize object-level perturbations to elicit backdoor behavior~\cite{shen2023django,cheng2024odscan}. Both directions face practical limitations. The test-time approach of~\cite{zhang2025test} relies on two independent and expensive search procedures, one for RMA and another for ODA, each requiring multiple forward passes per input. Even assuming perfect detection of attacked objects, running both procedures on every input would incur prohibitive computational cost. The trigger-inversion methods of~\cite{shen2023django,cheng2024odscan} assume the adversary commits to a specific victim class, an assumption violated by attacks such as BadDet, BadDet+, Morph, UBA, and Align, which train the backdoor to affect arbitrary classes. Methods that explicitly \emph{mitigate} backdoors in compromised detectors are far scarcer. To the best of our knowledge, \cite{zhang2024towards} proposes the only such mitigation method for object detection, but it is tailored specifically to Faster R-CNN and requires access to both clean and backdoor data.

A natural fallback is to reuse classification-oriented defenses, which are dominated by two paradigms: fine-tuning-based methods that retrain or regularise the model on a small clean subset, and pruning-based methods that identify and remove parameters most implicated in the backdoor~\citep{wu2022backdoorbench, dunnett2024countering}. Two features prevalent across this literature obstruct direct transfer to detection. First, many state-of-the-art methods in both families rely on an adversarial-example generation step that perturbs an input to elicit the backdoor's predicted class~\citep{wei2023sau,zhu2023enhancing,zeng2021adversarial,chai2022one}; this presumes a single-prediction, single-label abstraction that detection violates as each image produces many joint localisation-classification predictions, and under ODA the objective is the \emph{absence} of class expression rather than misclassification. Second, pruning-based methods predominantly assume a channel-level structure tied to CNN classifiers and do not extend cleanly to transformer-based detectors. These limitations motivate detector-aware mitigation strategies that provide (i) adversarial generation that addresses both RMA and ODA without assuming which the adversary is using, and (ii) a concentrated repair objective.

% \textbf{Adversarial Examples for Object Detection:}
% Generating adversarial examples for detection is intrinsically more complex than for classification since attacks can target different parts of the output, including classification scores, localization, and even the existence of detections~\cite{mi2023adversarial,nguyen2025survey}. While prior work has explored detection adversaries that include misclassification and disappearance, using these techniques for backdoor mitigation requires care, as a defender typically cannot assume the attacker’s objective (RMA vs.\ ODA) and must avoid over-engineering the mitigation procedure to a single failure mode. Building on these insights, we generate object-level adversarial examples that expose backdoor behavior without presuming any specific attacker objective, and by using them in a detection-aware fine-tuning procedure, analogous in spirit to fine-tuning-based backdoor mitigation for image classification (e.g.,~\cite{}).

\begin{figure}
    \centering
    \includegraphics[width=\linewidth]{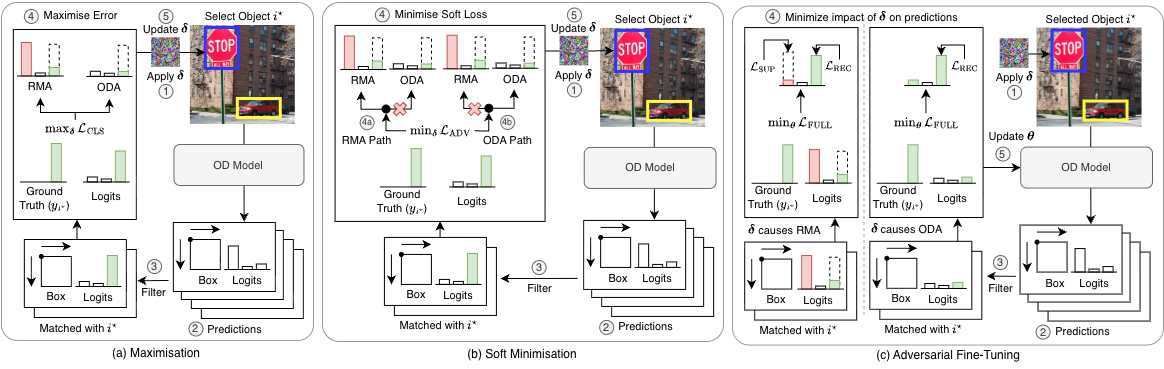}
    \caption{Overview of the proposed adversarial fine-tuning approach. (a) and (b) Adversarial example generation for a target object using the proposed maximisation and soft minimisation strategies. (c) Detection-aware fine-tuning objective used to mitigate the induced RMA- or ODA-style behaviour.} \vspace{-4mm}%How an adversarial example targeting object $i^{\star}$ in the image is found using the proposed (a) Maximisation and (b) Soft Minimisation methods. (c) How the model is updated to mitigate the misclassification, either RMA or ODA, induced by (a) or (b).}
    \label{fig:adv_framework} 
\end{figure}

\section{Preliminaries} \label{sec:prelim}

Let $f_\theta$ denote an object detector with parameters $\theta$. For an image $x$, let $\mathcal{B}=\{(\mathbf{b}_i, y_i)\}_{i=1}^{N}$ be the ground-truth set, where $\mathbf{b}_i \in \mathbb{R}^4$ and $y_i$ are the bounding-box coordinates and class label of the $i$-th object, respectively. The detector produces $\hat{N}$ candidate predictions $\hat{\mathcal{B}}=\{(\hat{\mathbf{b}}_j, \mathbf{z}_j)\}_{j=1}^{\hat{N}}$, where $\hat{\mathbf{b}}_j \in \mathbb{R}^4$ are predicted bounding boxes and $\mathbf{z}_j \in \mathbb{R}^{C}$ are class logits over the $C$ foreground classes. The corresponding inference-time class scores are denoted by $\mathbf{s}_j \in [0,1]^C$ where $s_{j,c}$ is the score for class $c$. Final detections are obtained by applying the detector's standard post-processing to $\{(\hat{\mathbf{b}}_j,\mathbf{s}_j)\}_{j=1}^{\hat{N}}$, including score thresholding at $\tau$ and non-maximum suppression (NMS).

During training, the detector's matching strategy $\mathcal{M}$ assigns predictions to ground-truth objects, yielding a mapping $\pi: \{1, \dots, \hat{N}\} \rightarrow \{0\} \cup \{1, \dots, N\}$. Here $\pi(j) = i > 0$ indicates that prediction $j$ is matched to ground-truth object $i$, while $\pi(j) = 0$ marks an unmatched (background) prediction. The predictions are then supervised by a detection loss that we abstract as
\begin{equation}
    \mathcal{L}_{\mathrm{OD}} = \mathcal{L}_{\mathrm{LOC}} + \mathcal{L}_{\mathrm{CLS}},
\end{equation}
where $\mathcal{L}_{\mathrm{LOC}}$ penalizes bounding-box regression error between each matched prediction $\hat{\mathbf{b}}_j$ and its assigned ground-truth box $\mathbf{b}_{\pi(j)}$, and $\mathcal{L}_{\mathrm{CLS}}$ penalizes classification error between the predicted logits $\mathbf{z}_j$ and the target label $y_{\pi(j)}$. This two-term decomposition is a simplification, as specific detector architectures may include additional components.

\textbf{Object-localized perturbation:}
Given a target object $i^\star$, we optimize an $\ell_\infty$-bounded perturbation $\delta$ restricted to the target box via a binary mask $\mathbf{M}(\mathbf{b}_{i^\star})$:
\begin{equation}
    x'=\Pi_{[0,1]}\!\big(x+\mathbf{M}(\mathbf{b}_{i^\star})\odot\delta\big),
    \quad \|\delta\|_\infty \le \epsilon.
\end{equation}

\textbf{Matched predictions for the target:} 
Let $\pi_{x'}$ denote the assignment computed on the perturbed image $x'$ using the matching strategy $\mathcal{M}$. We define the set of predictions matched to the target object as
\begin{equation}
    \mathcal{J}_{i^\star}(x')=\{j \mid \pi_{x'}(j)=i^\star\},\label{J_i}
\end{equation}
which is typically a singleton for set-based detectors and may contain multiple anchors or spatial locations for dense detectors. In Appendix~\ref{app:theory_restriction}, we provide a gradient-alignment rationale for restricting the inner optimisation to this target-matched set.

\textbf{Threat Model:}
%We assume the stronger threat model typically adopted within state-of-the-art classification-based backdoor mitigation:
We consider the standard repair setting used in backdoor mitigation: the defender has access only to a compromised detector and a limited set of clean data. We assume that the attack is object-specific, namely an ODA or RMA, but that the defender does not know the attack objective. These are object-targeting attacks in the sense that they affect predictions for trigger-bearing objects, but class-untargeted in that the attack is not tied to a specific victim class. Accordingly, we do not consider attacks that target background regions within existing images, such as the object-generation attack of~\cite{chan2022baddet}. Therefore, the defender's goal is to mitigate the unknown backdoor behaviour using only clean data. 

\textbf{Fine-tuning in Classification:}
Since the defender typically lacks prior knowledge of the trigger or attacker objective, adversarial fine-tuning has emerged as a prominent mitigation paradigm in image classification~\cite{wei2023sau,zhu2023enhancing,zeng2021adversarial}, using adversarial examples as \emph{surrogate backdoor data}. The approach is formulated as a minimax problem:
\begin{equation}
    \min_{\theta} \max_{\|\delta\|_p \le \epsilon} \mathcal{L}^{\mathrm{CE}}(f_{\theta}(x + \delta), y),
\end{equation}
where the inner maximization seeks the most harmful perturbation $\delta$ within an $\ell_p$-bounded neighborhood of input $x$ with ground-truth label $y$. Because a backdoored model relies on trigger-activated shortcut features, this inner step is biased toward perturbations that exploit related vulnerable pathways; the outer minimization then fine-tunes $\theta$ to remain correct under such perturbations, weakening the shortcut behavior and thereby mitigating the backdoor.

\section{Proposed Approach}\label{sec:proposed}

Building on the need for adversarial generation that handles an unknown attack objective and a concentrated repair objective identified in Section~\ref{sec:intro}, we propose an adversarial fine-tuning approach for backdoor mitigation in object detection that has two components. First, we reformulate the conventional maximisation-based adversarial example generation used in classification-oriented defences and introduce a complementary soft-branch minimisation alternative. Second, we redesign the outer fine-tuning objective to account for the multi-prediction structure of object detection models. The resulting fine-tuning objective is compatible with both adversarial generation strategies, and is summarised in Fig.~\ref{fig:adv_framework}.

% We propose an adversarial fine-tuning approach for backdoor mitigation in object detection that has two components. First, we reformulate the conventional maximisation-based adversarial example generation used in classification-oriented defences and introduce a complementary soft-branch minimisation alternative. Second, we redesign the outer fine-tuning objective to account for the multi-prediction structure of object detection models. 
% The resulting fine-tuning objective is compatible with both adversarial generation strategies. We summarise the overall framework in Fig.~\ref{fig:adv_framework}.

\subsection{Adversarial Example Generation}\label{sec:advgen}

We consider two detection-aware adversarial generation strategies: classification-loss maximisation (CLM), which adapts the standard adversarial maximisation paradigm to matched detector predictions, and soft-branch minimisation (SBM), which softly selects between RMA- and ODA-oriented objectives.

\subsubsection{Classification-Loss Maximisation}\label{sec:max}
As discussed in Section~\ref{sec:prelim}, loss maximisation is a standard strategy in classification-based backdoor mitigation for generating surrogate backdoor data. However, in object detection, naively maximising the full detection loss, $\mathcal{L}_{\mathrm{OD}}$, is sub-optimal for two reasons: it unnecessarily perturbs localisation and it does not isolate the effect of $\delta$ to the predictions relevant to the target object $i^\star$. To address this, we adopt a detection-aware formulation that focuses on $\mathcal{L}_{\mathrm{CLS}}$ and restricts the influence of $\delta$ to the relevant prediction set $\mathcal{J}_{i^\star}$ (see Fig.~\ref{fig:adv_framework} (a)). In Appendix~\ref{sec:appendix-flm-vs-clm}, we provide an ablation showing the impact of using the full detector loss $\mathcal{L}_{\mathrm{OD}}$. Therefore, we define the CLM strategy as
\begin{equation}\label{equ:clm}
    \max_{\delta} \frac{1}{|\mathcal{J}_{i^\star}(x')|} \sum_{j \in \mathcal{J}_{i^\star}(x')} \mathcal{L}_\text{CLS}(j).
\end{equation}
Restricting the adversarial loss to $\mathcal{J}_{i^\star}(x')$ concentrates the optimisation on predictions matched to the target object, hence reducing interference and gradient noise from unrelated objects or background proposals. However, this strategy does not explicitly distinguish between the objectives of RMA and ODA. In particular, maximising $\mathcal{L}_{\mathrm{CLS}}$ is only an indirect surrogate for either inducing a dominant incorrect class or suppressing foreground detections. This motivates introducing an alternative objective that more directly reflects these two behaviours. In Appendix~\ref{app:theory_misalignment}, we provide a supporting score-level analysis.

\subsubsection{Soft-Branch Minimisation}\label{sec:smin}

As an alternative, we introduce SBM (see Fig.~\ref{fig:adv_framework} (b)). SBM accommodates an unknown attack objective by defining two optimisation branches (i.e., 4a and 4b in Fig.~\ref{fig:adv_framework} (b)), corresponding to RMA and ODA, and combining them through a soft gate that favours the branch that is easier to satisfy. The intuition is that the resulting perturbation $\delta$ is biased toward the branch that better aligns with the detector's underlying backdoor behaviour, without requiring the defender to know the attack objective.

For each matched prediction $j \in \mathcal{J}_{i^\star}(x')$, we define three quantities from the class scores $s_{j,c}(x')$:
\begin{equation}
    s^{\mathrm{gt}}_j = s_{j,y_{i^\star}}(x'), \qquad s^{\mathrm{oth}}_j = \max_{c \neq y_{i^\star}} s_{j,c}(x'), \qquad s^{\mathrm{max}}_j = \max_{c} s_{j,c}(x'),
\end{equation}
which denote the ground-truth class score, the highest non-target class score, and the maximum foreground score, respectively.

\textbf{Misclassification (RMA):}
To encourage a false class to exceed the detection threshold $\tau$, we define
\begin{equation}
    \mathcal{L}_{\mathrm{RMA}}(j) = \mathrm{softplus}(\tau - s^{\mathrm{oth}}_j),
\end{equation}
where $\mathrm{softplus}(t) = \log(1 + e^t)$ ensures non-zero gradients everywhere. However, minimising this term alone may spread score mass across several incorrect classes, much like the maximisation objective above. To encourage a single dominant incorrect class, we add an entropy-based concentration penalty over the normalised non-target scores $\bar{\mathbf{s}}_{j,\neg y}$:
\begin{equation}
    \mathcal{L}_{\mathrm{CON}}(j) = -\sum_{c \neq y_{i^\star}} \bar{s}_{j,c} \log(\bar{s}_{j,c} + \zeta),
\end{equation}
where $\zeta > 0$ is a small stabilising constant. Minimising $\mathcal{L}_{\mathrm{CON}}(j)$ encourages the non-target score distribution to concentrate on a single incorrect class. In Appendix~\ref{app:theory_softmin}, we show that the proposed soft minimisation objective can be interpreted as an entropy-regularised branch selector between the RMA and ODA paths, while in Appendix~\ref{app:theory_concentration}, we clarify the role of $\mathcal{L}_{\mathrm{CON}}$.

\textbf{Disappearance (ODA):}
To suppress all foreground detections, we define
\begin{equation}
    \mathcal{L}_{\mathrm{ODA}}(j) = \mathrm{softplus}(s^{\mathrm{max}}_j - \tau),
\end{equation}
which encourages all class scores below $\tau$, making the target object less likely to survive score filtering.

\textbf{Joint objective:}
Since the attack objective is unknown, we combine both branches using a soft gate that prioritises the objective that is closer to satisfaction: $g^{\mathrm{RMA}}_j,\; g^{\mathrm{ODA}}_j = \mathrm{softmax}\left[-\mathcal{L}_{\mathrm{RMA}}(j),\; -\mathcal{L}_{\mathrm{ODA}}(j)\right]$.
The resulting per-prediction adversarial loss is $\mathcal{L}_{\mathrm{ADV}}(j) = g^{\mathrm{RMA}}_j \left[\mathcal{L}_{\mathrm{RMA}}(j) + \mathcal{L}_{\mathrm{CON}}(j)\right] + g^{\mathrm{ODA}}_j \mathcal{L}_{\mathrm{ODA}}(j)$, 
and the full soft minimisation objective, averaged over matched predictions, is
\begin{equation}\label{equ:sbm}
    \min_{\delta} \; \frac{1}{|\mathcal{J}_{i^\star}(x')|} \sum_{j \in \mathcal{J}_{i^\star}(x')} \mathcal{L}_{\mathrm{ADV}}(j).
\end{equation}
We solve this optimization problem using $K$-step projected gradient descent with sign-gradient updates and projection. As in the maximisation formulation, restricting optimisation to $\mathcal{J}_{i^\star}(x')$ focuses the gradients on matched predictions. This soft gate provides a differentiable mixture of the RMA and ODA objectives via softmax-normalized negative losses. The auxiliary concentration term, $\mathcal{L}^{\mathrm{CON}}_j$, is branch-specific and shapes the RMA path by encouraging a single dominant incorrect class.

\subsection{Adversarial Fine-Tuning}\label{sec:advft}
Given an adversarially perturbed image $x'$ produced by either method in Section~\ref{sec:advgen}, we fine-tune $\theta$ using the standard detection loss $\mathcal{L}_{\mathrm{OD}}$ together with a dual-objective defence loss computed directly on the target-matched predictions $\mathcal{J}_{i^\star}(x')$ (see Fig.~\ref{fig:adv_framework} (c)). Because $\mathcal{L}_{\mathrm{OD}}$ aggregates over all candidate predictions, the repair signal at $i^\star$ is otherwise diluted by background and non-target contributions; restricting the defence loss to $\mathcal{J}_{i^\star}(x')$ concentrates this signal on the predictions that need to have their classification restored. This loss comprises two terms that explicitly counteract the score-level events defining the two attack modes: a recovery term that encourages the ground-truth class score to exceed $\tau$, and a suppression term that drives competing class scores below $\tau$.

\textbf{Recovery:}
We encourage the ground-truth class score to recover a margin above the detection threshold via
\begin{equation}
    \mathcal{L}_{\mathrm{REC}}(j) = \mathrm{softplus}(\tau - s^{\mathrm{gt}}_j).
\end{equation}
Minimising this term counters ODA and RMA induced backdoors by increasing the target object's ground-truth score on the perturbed input, encouraging at least one matched prediction to survive score filtering (see Fig.~\ref{fig:adv_framework} (c) 4).

\textbf{Suppression:}
We penalise high non-target activations through
\begin{equation}
    \mathcal{L}_{\mathrm{SUP}}(j) = \mathrm{softplus}(s^{\mathrm{oth}}_j - \tau).
\end{equation}
Minimising this term prevents a degenerate solution in which the detector restores the ground-truth score but simultaneously assigns large scores to competing classes, which would otherwise lead to misclassification after thresholding. This safeguard is particularly relevant when $\delta$ induces RMA-aligned behaviour, as depicted by the '$\delta$ causes ODA' path in Fig.~\ref{fig:adv_framework}(c).

We combine $\mathcal{L}_{\mathrm{OD}}$ with the defence loss $\mathcal{L}_{\mathrm{DEF}}(j) = \mathcal{L}_{\mathrm{REC}}(j) + \mathcal{L}_{\mathrm{SUP}}(j)$, over target-matched predictions as
\begin{equation}
    \mathcal{L}_{\mathrm{FULL}}(\theta) = \mathcal{L}_{\mathrm{OD}}(x;\theta) + \mathcal{L}_{\mathrm{OD}}(x'; \theta) + \frac{\lambda}{|\mathcal{J}_{i^\star}(x')|} \sum_{j \in \mathcal{J}_{i^\star}(x')} \mathcal{L}_{\mathrm{DEF}}(j).
\end{equation}

Restricting $\mathcal{L}_{\mathrm{DEF}}$ to $\mathcal{J}_{i^\star}(x')$ mirrors the target-matched restriction used during adversarial generation in Section~\ref{sec:advgen}, ensuring that both the inner and outer loops concentrate on the same set of predictions.

\subsection{Optimisation Framework}
\label{sec:overall_opt}

Our defence alternates between (i) generating object-localised adversarial examples, using either Fig.~\ref{fig:adv_framework} (a) or (b), and (ii) fine-tuning the compromised detector to restore correct classifications while suppressing spurious activations, see Fig.~\ref{fig:adv_framework} (c). This yields a bilevel-style procedure in which an inner loop approximately solves for $\delta^\star$ and an outer loop updates $\theta$.

\textbf{Target object selection:}
Since each input may contain multiple objects, adversarial generation requires selecting a target $(\mathbf{b}_{i^\star}, y_{i^\star}) \in \mathcal{B}$. We consider two selection strategies:
\begin{enumerate}[leftmargin=14pt,noitemsep]
    \item \emph{Random selection (RS):} sample $i^\star$ uniformly from $\{1,\dots,N\}$.
    \item \emph{Filtered weighted selection (FWS):} restrict candidates to objects that are correctly detected on the clean input (matched prediction IoU above the relevant threshold, with no high-scoring overlapping incorrect predictions), and then sample proportionally to their clean inference confidence.
\end{enumerate}
RS provides unbiased coverage across objects, whereas FWS prioritises objects that are already cleanly detected, hence avoiding adversarial updates that merely exploit pre-existing detection ambiguity. In Appendix~\ref{app:theory_fws}, we provide an expected-informativeness rationale for FWS.

\textbf{Inner loop:}
Given a clean image $x$ and target index $i^\star$, we form $x'(\delta)=\Pi_{[0,1]}(x+\mathbf{M}(\mathbf{b}_{i^\star})\odot\delta)$. Depending on the adversarial generation strategy, the inner problem either solves Equation~\ref{equ:clm} or ~\ref{equ:sbm} in the CLM or SBM cases respectively.
% \begin{equation}
% \delta^\star \in
% \arg\min_{\|\delta\|_\infty \le \epsilon}
% \frac{1}{|\mathcal{J}_{i^\star}(x'(\delta))|}
% \sum_{j \in \mathcal{J}_{i^\star}(x'(\delta))}
% \mathcal{L}_{\mathrm{ADV}}(j),
% \end{equation}
% for soft minimisation, and
% \begin{equation}
% \delta^\star \in
% \arg\max_{\|\delta\|_\infty \le \epsilon}
% \frac{1}{|\mathcal{J}_{i^\star}(x'(\delta))|}
% \sum_{j \in \mathcal{J}_{i^\star}(x'(\delta))}
% \mathcal{L}_{\mathrm{CLS}}(j),
% \end{equation}
% for maximisation.
% %corresponding to the two adversarial generation strategies introduced in Section~\ref{sec:advgen}.

\textbf{Outer loop:}
Given the perturbed image $x'$, we minimise $\mathcal{L}_{\mathrm{FULL}}(\theta)$ with respect to $\theta$. This jointly optimises the standard detection objective on both $x$ and $x'$ while applying the targeted recovery and suppression terms to the matched predictions $\mathcal{J}_{i^\star}(x')$, thereby steering the detector toward correct classification of the target object under adversarial perturbation.

\section{Evaluation} \label{sec:results}

In this section, we evaluate the proposed method against standard fine-tuning and pruning, adapted from image-classification backdoor mitigation~\cite{liu2018fine}. Following the evaluation methodology of BadDet+~\cite{dunnett2026baddet+}, we conduct a comprehensive study across diverse settings, covering four datasets and three detector architectures. We make our benchmarking framework publicly available on GitHub\footnote{The code is included with the submission and will be released upon acceptance.}.

%\subsection{Experimental Setup} 

We evaluate untargeted ODA and RMA. For ODA we use BadDet+, the only attack reliably inducing object disappearance; for RMA we use both BadDet and BadDet+, which achieve high ASR but differ in retaining the original-class detection alongside the target. Experiments are conducted on Pascal VOC~\cite{everingham2010pascal} and MTSD~\cite{ertler2020mapillary} with DINO~\cite{zhang2022dino}, FCOS~\cite{tian2019fcos}, and YOLOv5m6~\cite{yolov5}; COCO ablations~\cite{lin2014microsoft} appear in Appendix~\ref{sec:appendix-coco}, and real-world transfer of MTSD-trained mitigations to PTSD~\cite{doan2024credibility} in Appendix~\ref{sec:appendix-ptsd}. For MTSD, we use traffic-sign meta-class labels and exclude \emph{other-sign} to alleviate class imbalance.

Triggers are placed randomly for MTSD/PTSD and centred in bounding boxes for Pascal VOC, where high object density makes random placement impractical. We use a blue trigger for PTSD compatibility, with alternatives explored in Appendix~\ref{appendix:triggers}. Our main experiments use $\sim$5\% clean training data (500 samples for Pascal VOC, 100 for MTSD) across five random splits, with 2.5\% and 1.25\% additionally tested. We compare CLM and SBM against FT, FP, and FP+FT~\cite{liu2018fine}, and FT-SAM~\cite{zhu2023enhancing}, evaluating RS and FWS strategies (Section~\ref{sec:overall_opt}) with and without the defence terms. Full implementation details, hyperparameters, and additional results are in Appendices~\ref{sec:appendix-selection-hyper}, \ref{sec:appendix-additional-ablations}, and \ref{sec:appendix-additional-results}. We evaluate each method using three metrics: ratio mAP (RmAP), post-mitigation clean mAP divided by the pre-mitigation value, as well as ASR, and TDR reported using IoU 0.5. An RMA attack succeeds when the target class is positively detected; an ODA attack succeeds when no positive detection of the original class occurs. TDR is therefore omitted for ODA since $\text{ASR} = 1 - \text{TDR}$.

\begin{figure}[t]
    \centering
    \begin{subfigure}[b]{0.535\textwidth}
        \centering
        \includegraphics[width=\linewidth]{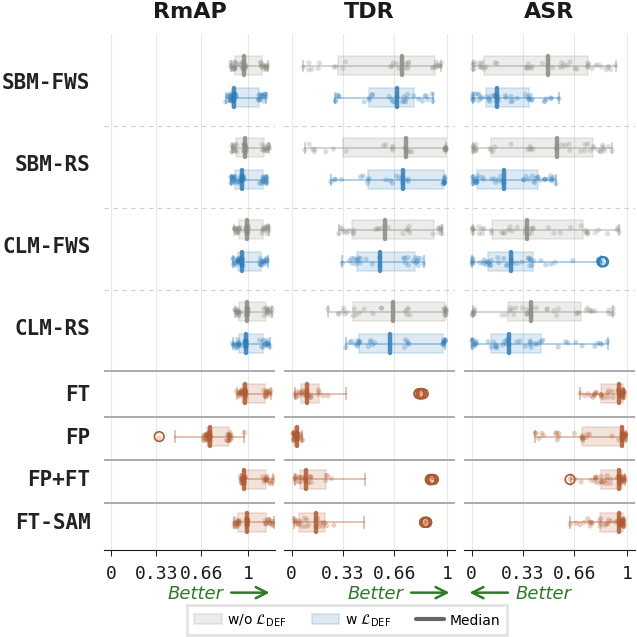}
        \caption{Pascal VOC}
    \end{subfigure}
    \begin{subfigure}[b]{0.45\textwidth}
        \centering
        \includegraphics[width=\linewidth]{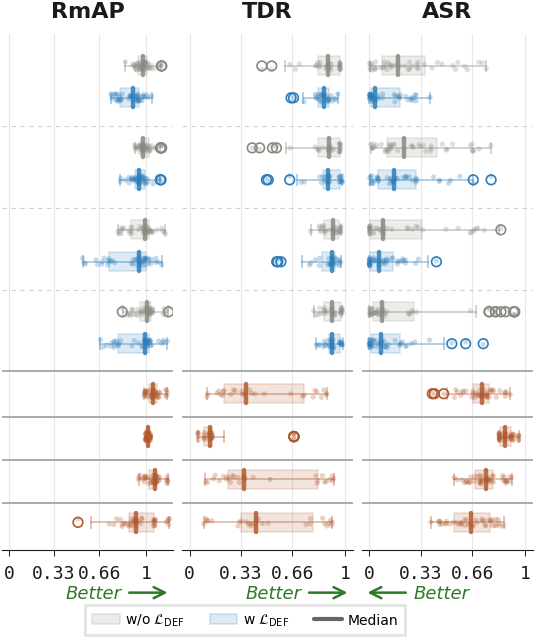}
        \caption{MTSD}
    \end{subfigure}
\caption{Performance comparison of four adversarial fine-tuning variants (SBM-FWS, SBM-RS, CLM-FWS, CLM-RS) alongside four baseline defenses (FT, FP, FP+FT, FT-SAM).}
\label{fig:overall_results}
\end{figure}

\begin{figure}[t]
    \centering
    % Adjust the 'height=3.5cm' value across all images until they fit nicely within the margins
    \subcaptionbox{Pascal VOC RMA}{%
        \includegraphics[height=3.9cm]{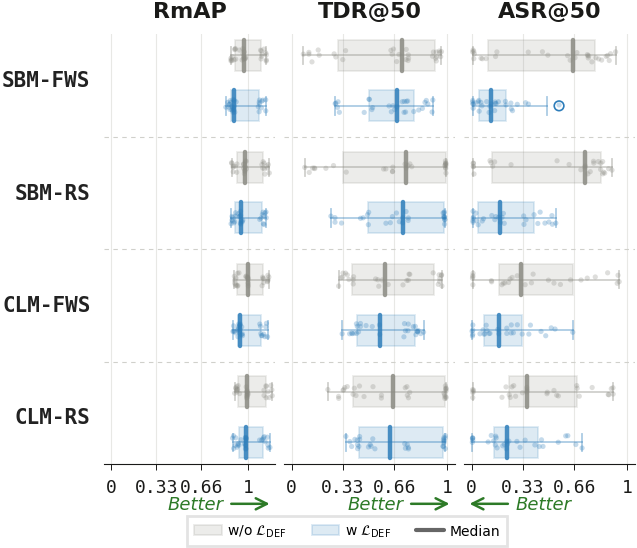}%
    }\hfill
    \subcaptionbox{MTSD RMA}{%
        \includegraphics[height=3.9cm]{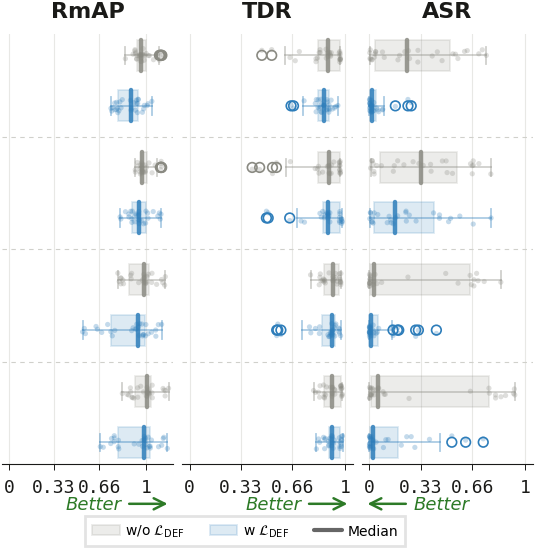}%
    }\hfill
    \subcaptionbox{Pascal VOC ODA}{%
        \includegraphics[height=3.9cm]{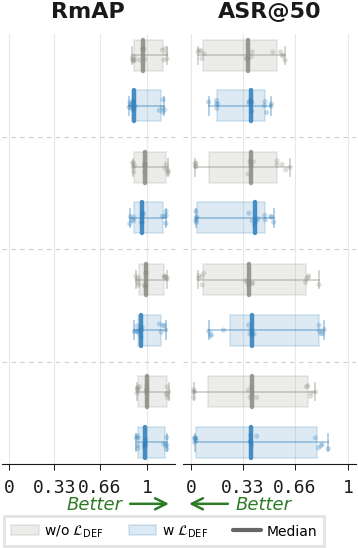}%
    }\hfill
    \subcaptionbox{MTSD ODA}{%
        \includegraphics[height=3.9cm]{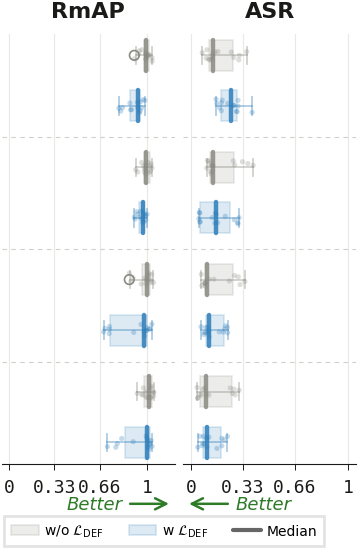}%
    }
    \caption{Performance comparison of four adversarial fine-tuning variants (SBM-FWS, SBM-RS, CLM-FWS, CLM-RS) seperated by attack objective.} \vspace{-6mm}
    \label{fig:attack_type}
\end{figure}

\begin{figure}[t]
    \centering
    \begin{subfigure}[b]{0.5\textwidth}
        \centering
        \includegraphics[width=\linewidth]{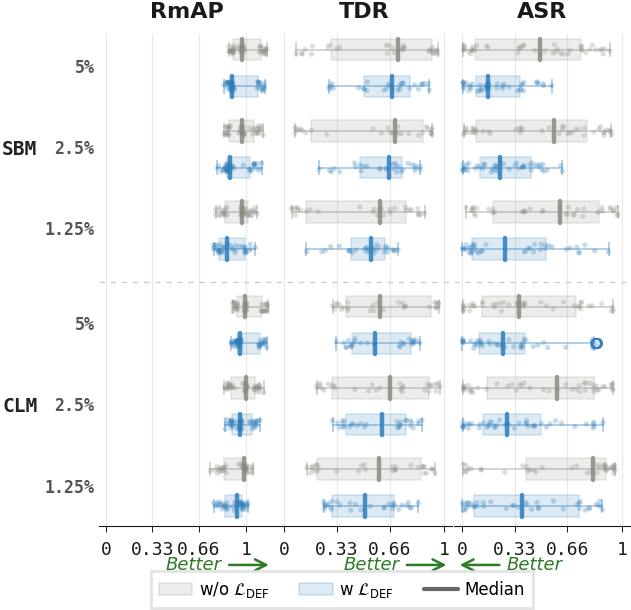}
        \caption{Pascal VOC}
    \end{subfigure}
    \begin{subfigure}[b]{0.425\textwidth}
        \centering
        \includegraphics[width=\linewidth]{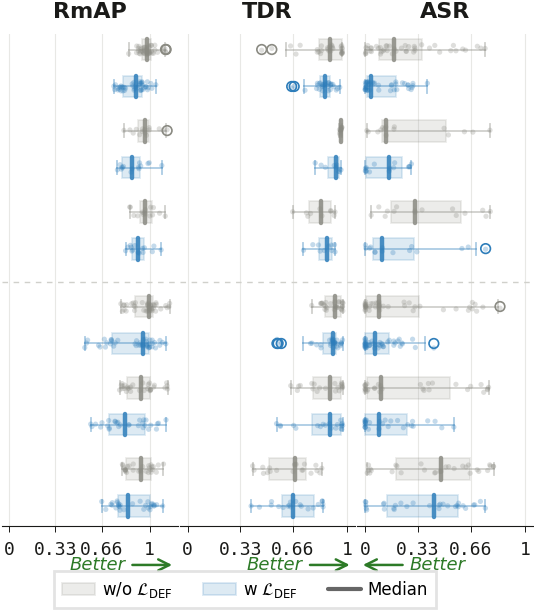}
        \caption{MTSD}
    \end{subfigure}
\caption{Performance comparison of SBM-FWS and CLM-FWS separated by attack objective.} \vspace{-5mm}
\label{fig:num_samples}
\end{figure}

\textbf{Main Results:} Figure~\ref{fig:overall_results} summarises performance with 5\% clean training data, with Appendix Tables~\ref{tab:defense_boxplot_medians_overal_voc} and~\ref{tab:defense_boxplot_medians_overall_mtsd} reporting the median performance. Compared to FT, FP, FP+FT, and FT-SAM, the proposed CLM and SBM reduce ASR and raise TDR, at a modest cost to RmAP. The strongest configuration, SBM-FWS with $\mathcal{L}_{\mathrm{DEF}}$, reaches median (ASR, TDR) of $(0.163, 0.676)$ on VOC and $(0.037, 0.861)$ on MTSD, against the best baselines at $(0.945, 0.156)$ and $(0.653, 0.427)$ respectively, at a RmAP cost of at most $0.10$. Adding $\mathcal{L}_{\mathrm{DEF}}$ reduces median ASR by 67\% for SBM-FWS and 29\% for CLM-FWS on VOC, and by 80\% and 33\% respectively on MTSD, while reducing TDR variance with some additional RmAP cost. This is consistent with the motivation in Section~\ref{sec:advft}: applying the repair signal directly to $\mathcal{J}_{i^\star}(x')$ avoids the dilution that target-object gradients undergo within the aggregated $\mathcal{L}_{\mathrm{OD}}$. Similarly, FWS lowers ASR relative to RS. For SBM with $\mathcal{L}_{\mathrm{DEF}}$, median ASR drops by a further 21\% on VOC and 76\% on MTSD, while the effect is smaller and mixed for CLM. This comes at the cost of a small reduction in RmAP and occasionally TDR, reflecting that restricting candidates to cleanly-detected objects reduces the probability that the inner loop exploits pre-existing detection ambiguity rather than backdoor behaviour (Appendix~\ref{app:theory_fws}). Comparing the FWS variants directly, SBM-FWS improves median ASR and TDR by roughly 36\% and 20\% relative to CLM-FWS on VOC, while on MTSD the two achieve similar ASR and TDR. This reflects the design distinction in Section~\ref{sec:advgen}: CLM maximises $\mathcal{L}_{\mathrm{CLS}}$ over $\mathcal{J}_{i^\star}(x')$, an indirect surrogate for the score-threshold events characterising RMA and ODA, whereas SBM optimises $\mathcal{L}_{\mathrm{RMA}}$ and $\mathcal{L}_{\mathrm{ODA}}$ directly via a soft gate favouring the branch closer to satisfaction (Appendix~\ref{app:theory_misalignment}).

\textbf{Attack Objective:}
Figure~\ref{fig:attack_type} reports separate results for ODA and RMA, with Appendix Tables~\ref{tab:defense_attack_boxplot_medians_voc},~\ref{tab:defense_attack_boxplot_medians_mtsd} reporting the median performance. The proposed defense terms are particularly beneficial against RMA, reducing median ASR by roughly 80\% on VOC ($0.649 \rightarrow 0.126$) and 94\% on MTSD ($0.239 \rightarrow 0.014$) for SBM-FWS, while also decreasing variability in both ASR and TDR. The ODA results are more nuanced. On VOC, adding $\mathcal{L}_{\mathrm{DEF}}$ barely changes ODA ASR for SBM-FWS ($0.364 \rightarrow 0.379$), while on MTSD it raises ODA ASR ($0.138 \rightarrow 0.251$). Because the defender does not know the attack objective, the relevant criterion is robustness across both attack objectives, under which SBM-FWS with the proposed defense terms is the most reliable choice. If ODA were known in advance, the preferred configuration would be dataset-dependent: SBM-RS on the more complex Pascal VOC, and CLM-FWS on MTSD.

\textbf{Data Availability:}
Figure~\ref{fig:num_samples} illustrates the impact of decreasing the amount of clean data for SBM-FWS and CLM-FWS, with Appendix Tables~\ref{tab:num_samples_boxplot_medians_voc},~\ref{tab:num_samples_boxplot_medians_mtsd} reporting the median performance. Even with the clean-data budget halved or quartered, the proposed defense remains highly effective: as the budget shrinks from 5\% to 1.25\%, median ASR for SBM with $\mathcal{L}_{\mathrm{DEF}}$ rises only from $0.163$ to $0.269$ on VOC and from $0.037$ to $0.108$ on MTSD. Notably, these results are still below the strongest baseline ASR ($0.945$ on VOC, $0.653$ on MTSD) measured at the full 5\% budget. Variance in TDR and ASR does grow as clean data shrinks; however, SBM with $\mathcal{L}_{\mathrm{DEF}}$ is more insulated from these median and variance changes than CLM.

\textbf{Key Findings:}
Taken together, the proposed adversarial fine-tuning framework delivers: (i) median ASR reductions of 83\% (VOC) and 94\% (MTSD) over the strongest baseline for SBM-FWS with $\mathcal{L}_{\mathrm{DEF}}$, with TDR gains of $0.52$ and $0.43$ respectively at a RmAP cost of at most $0.10$; (ii) $\mathcal{L}_{\mathrm{DEF}}$ provides a major contribution, reducing median ASR by 67--80\% for SBM-FWS and rising to 80--94\% on the RMA attack subset; (iii) further ASR reductions from FWS over RS, most strikingly for SBM with $\mathcal{L}_{\mathrm{DEF}}$ (21\% on VOC, 76\% on MTSD), with a smaller and mixed effect for CLM; and (iv) SBM-FWS with $\mathcal{L}_{\mathrm{DEF}}$ is more insulated from data scarcity than CLM, still outperforming the strongest baseline (measured at 5\% clean data) when fine-tuned on as little as 1.25\% clean data.

\section{Conclusion and Future Work}
We proposed a detection-aware adversarial fine-tuning method for mitigating object-level backdoors under a practical repair setting, where the defender has access only to a compromised detector and a small clean dataset. Our approach addresses two structural mismatches between classification and detection settings: (i) adversarial generation must handle an unknown attack objective, and (ii) the repair signal must concentrate on the targeted object. We address the first by restricting adversarial generation to target-matched predictions through two strategies, CLM and SBM, the latter softly combining RMA- and ODA-aligned objectives to account for the attack objective being unknown. We address the second through a dual-objective defence loss applied directly to the target-matched predictions during fine-tuning. Across CNN- and Transformer-based detectors and multiple datasets, this combination improves the ASR-TDR trade-off relative to classification-oriented baselines while maintaining a favorable trade-off with clean detection performance.

Because the defender does not know whether the compromised model is affected by RMA or ODA, an effective mitigation strategy must perform reliably across both. Our results indicate that SBM with the proposed defence terms, particularly when paired with filtered weighted selection of target objects (FWS, Section~\ref{sec:overall_opt}), provides the most robust overall trade-off. This reflects the closer alignment of SBM's objectives with the events characterising RMA and ODA, and is also more stable than CLM in low-data regimes. Nevertheless, fully recovering correct detections for trigger-bearing objects remains challenging, consistent with limitations observed for fine-tuning-based defences in the classification setting~\cite{dunnett2024countering}, with TDR retaining a long lower tail in some settings.

Building on these findings, two directions for future work seem particularly promising. The first is developing mechanisms that more directly improve TDR while maintaining low ASR and competitive RmAP. Pruning-based strategies are a natural candidate; however, existing pruning defences are largely designed for CNN architectures, and extending them to Transformer-based detectors, in particular, how to prune or regularise attention mechanisms effectively, remains open. The second is broadening mitigation beyond object-centric attacks. Object-generation attacks~\cite{chan2022baddet} manipulate background regions to induce phantom detections, a behaviour that is not directly addressed by our object-specific framework. Developing unified mitigation strategies that handle both localised triggers and background-induced false detections is an important next step.

\bibliography{main}
\bibliographystyle{plainnat}

\appendix

\section{Theoretical Analysis}
\label{app:theory}

We provide supporting analysis for the design choices in Sections~\ref{sec:advgen}--\ref{sec:overall_opt}. The goal is not to establish a full end-to-end robustness theorem for modern object detectors, which would require strong architecture- and post-processing-specific assumptions. Instead, we show that the proposed objectives are aligned with the score-level events that characterize object-level backdoor attacks, and that the outer defence terms induce a meaningful repaired margin against those events.

Throughout, we analyse the detector scores $\mathbf{s}_j(x') \in [0,1]^C$ associated with predictions matched to the target object. This abstraction is agnostic to whether the detector is anchor-based, anchor-free, or set-based, and to whether class scores arise from sigmoid, softmax, or detector-specific scoring heads. We assume that the training-time assignment mechanism yields a nonempty set $\mathcal{J}_{i^\star}(x')$ of predictions matched to the target object $i^\star$, as is standard for common dense and set-based assigners. If an implementation can produce an empty set, the definitions below can be interpreted using the usual practical fallback to the clean-image assignment. Unless stated otherwise, gradient statements are conditioned on the current matched set used to evaluate the loss and are taken away from score ties introduced by $\max$ operators; at ties, the same conclusions hold in the usual subgradient sense.

\subsection{Score-Level Margins and Attack Events}
\label{app:theory_margins_setup}

For each matched prediction $j \in \mathcal{J}_{i^\star}(x')$, recall that
\begin{equation}
    s^{\mathrm{gt}}_j = s_{j,y_{i^\star}}(x'), \quad
    s^{\mathrm{oth}}_j = \max_{c \neq y_{i^\star}} s_{j,c}(x'), \quad
    s^{\mathrm{max}}_j = \max_c s_{j,c}(x').
\end{equation}
We define the corresponding score margins as
\begin{equation}
    m^{\mathrm{RMA}}_j = s^{\mathrm{oth}}_j - \tau,
    \quad
    m^{\mathrm{ODA}}_j = \tau - s^{\mathrm{max}}_j,
    \quad
    m^{\mathrm{REC}}_j = s^{\mathrm{gt}}_j - \tau,
    \quad
    m^{\mathrm{SUP}}_j = \tau - s^{\mathrm{oth}}_j .
\end{equation}
These margins encode the score-threshold conditions associated with the attack and defence objectives:
\begin{itemize}[leftmargin=6mm,noitemsep]
    \item $m^{\mathrm{RMA}}_j \ge 0$ indicates that at least one non-ground-truth class score is at or above the detection threshold $\tau$;
    \item $m^{\mathrm{ODA}}_j \ge 0$ indicates that all foreground class scores are at most $\tau$;
    \item $m^{\mathrm{REC}}_j \ge 0$ indicates that the ground-truth class score is at or above $\tau$;
    \item $m^{\mathrm{SUP}}_j \ge 0$ indicates that all non-ground-truth class scores are at most $\tau$.
\end{itemize}
Because the defender does not know the attacker's target class in an RMA, we use the strongest non-ground-truth score $s_j^{\mathrm{oth}}$ as a defender-agnostic surrogate. This condition upper-bounds targeted RMA behaviour in the sense that, if any attacker-chosen target class crosses the threshold, then $s_j^{\mathrm{oth}}$ also crosses the threshold.

The proposed inner and outer losses are smooth surrogates of these margin conditions:
\begin{align}
    \mathcal{L}_{\mathrm{RMA}}(j)&=\mathrm{softplus}(-m^{\mathrm{RMA}}_j),\quad
    \mathcal{L}_{\mathrm{ODA}}(j)=\mathrm{softplus}(-m^{\mathrm{ODA}}_j),\\
    \mathcal{L}_{\mathrm{REC}}(j)&=\mathrm{softplus}(-m^{\mathrm{REC}}_j),\quad
    \mathcal{L}_{\mathrm{SUP}}(j)=\mathrm{softplus}(-m^{\mathrm{SUP}}_j).
\end{align}

\paragraph{Proposition A1 (order consistency of the score surrogates).}
For each of the four losses above, minimising the loss is equivalent to maximising the corresponding score margin in the sense that the ordering of candidate perturbations is preserved. In particular, for any two perturbed inputs $x'_1,x'_2$ and any matched prediction $j$,
\begin{align}
    m^{\mathrm{RMA}}_j(x'_1) > m^{\mathrm{RMA}}_j(x'_2)
    &\iff
    \mathcal{L}_{\mathrm{RMA}}(j;x'_1) <
    \mathcal{L}_{\mathrm{RMA}}(j;x'_2),\\
    m^{\mathrm{ODA}}_j(x'_1) > m^{\mathrm{ODA}}_j(x'_2)
    &\iff
    \mathcal{L}_{\mathrm{ODA}}(j;x'_1) <
    \mathcal{L}_{\mathrm{ODA}}(j;x'_2),
\end{align}
and the same equivalence holds for $\mathcal{L}_{\mathrm{REC}}$ and $\mathcal{L}_{\mathrm{SUP}}$ with their corresponding margins.

\paragraph{Proof.}
The function $\mathrm{softplus}(t)=\log(1+e^t)$ is strictly increasing in $t$. Hence $\mathrm{softplus}(-m)$ is strictly decreasing in $m$, which gives the stated ordering equivalences. \hfill $\square$

Proposition~A1 shows that the proposed losses preserve the score-threshold logic of the corresponding attack and defence events. They are monotone surrogates of the relevant margins, differentiable away from score ties and subdifferentiable at ties.

\subsection{Gradient-Alignment Rationale for Target-Matched Restriction}
\label{app:theory_restriction}

A central design choice in both classification-loss maximisation and soft branch minimisation is to restrict adversarial generation to predictions matched to the target object. This restriction can be understood through a gradient-alignment argument.

Let $L_{\mathrm{tar}}(\delta)$ denote the average loss over the target-matched set $\mathcal{J}_{i^\star}(x'(\delta))$, and let $L_{\mathrm{nui}}(\delta)$ collect the remaining prediction terms that would appear in an unrestricted image-level objective. Writing
\begin{equation}
    L_{\mathrm{all}}(\delta)=L_{\mathrm{tar}}(\delta)+L_{\mathrm{nui}}(\delta),
\end{equation}
the corresponding unrestricted gradient direction can be decomposed as
\begin{equation}
    g_{\mathrm{all}} = g_{\mathrm{tar}} + g_{\mathrm{nui}},
    \qquad
    g_{\mathrm{tar}} = \nabla_\delta L_{\mathrm{tar}},
    \qquad
    g_{\mathrm{nui}} = \nabla_\delta L_{\mathrm{nui}} .
\end{equation}

\paragraph{Proposition A2 (target restriction reduces gradient dilution).}
Assume $g_{\mathrm{tar}}\neq 0$ and $g_{\mathrm{all}}\neq 0$, and suppose there exist $\rho \in [0,1]$ and $\beta \ge 0$ such that
\begin{equation}
    |\langle g_{\mathrm{tar}}, g_{\mathrm{nui}} \rangle|
    \le
    \rho \|g_{\mathrm{tar}}\|^2,
    \qquad
    \|g_{\mathrm{nui}}\|
    \le
    \beta \|g_{\mathrm{tar}}\|.
\end{equation}
Then the cosine alignment between the unrestricted gradient and the target gradient satisfies
\begin{equation}
    \cos(g_{\mathrm{tar}}, g_{\mathrm{all}})
    \ge
    \frac{1-\rho}{1+\beta}.
\end{equation}
In contrast, using only $L_{\mathrm{tar}}$ yields perfect alignment with the target objective, since $\cos(g_{\mathrm{tar}},g_{\mathrm{tar}})=1$.

\paragraph{Proof.}
By definition,
\begin{equation}
    \cos(g_{\mathrm{tar}},g_{\mathrm{all}})
    =
    \frac{\langle g_{\mathrm{tar}}, g_{\mathrm{tar}} + g_{\mathrm{nui}} \rangle}
    {\|g_{\mathrm{tar}}\| \, \|g_{\mathrm{tar}} + g_{\mathrm{nui}}\|}.
\end{equation}
Using the assumed bound on the inner product,
\begin{align}
    \langle g_{\mathrm{tar}}, g_{\mathrm{tar}} + g_{\mathrm{nui}} \rangle
    &=
    \|g_{\mathrm{tar}}\|^2 + \langle g_{\mathrm{tar}}, g_{\mathrm{nui}} \rangle \nonumber\\
    &\ge
    (1-\rho)\|g_{\mathrm{tar}}\|^2 .
\end{align}
Using the triangle inequality and the norm bound on $g_{\mathrm{nui}}$,
\begin{align}
    \|g_{\mathrm{tar}} + g_{\mathrm{nui}}\|
    &\le
    \|g_{\mathrm{tar}}\| + \|g_{\mathrm{nui}}\| \nonumber\\
    &\le
    (1+\beta)\|g_{\mathrm{tar}}\|.
\end{align}
Combining these two inequalities gives
\begin{equation}
    \cos(g_{\mathrm{tar}},g_{\mathrm{all}})
    \ge
    \frac{(1-\rho)\|g_{\mathrm{tar}}\|^2}
    {(1+\beta)\|g_{\mathrm{tar}}\|^2}
    =
    \frac{1-\rho}{1+\beta}.
\end{equation}
\hfill $\square$

Proposition~A2 formalises the intuition that unrelated objects and background proposals can dilute the target-object gradient unless their gradients are sufficiently aligned with the target objective. Restricting optimisation to $\mathcal{J}_{i^\star}(x'(\delta))$ therefore improves the alignment of the inner-loop update with the intended target-object objective.

\subsection{Partial Alignment of Classification-Loss Maximisation}
\label{app:theory_misalignment}

The classification-loss maximisation baseline in Section~\ref{sec:advgen} provides a useful detection-adapted surrogate, but it does not directly optimise the score-threshold events that define RMA and ODA.

For detectors with independent sigmoid-based classification heads, the classification loss for a matched prediction decomposes into class-wise terms, possibly with focal, label-smoothing, or implementation-specific weights. The following BCE form illustrates the basic mechanism:
\begin{equation}
    \mathcal{L}_{\mathrm{CLS}}(j)
    =
    -\log \sigma(z_{j,y_{i^\star}})
    -
    \sum_{c\neq y_{i^\star}} \log(1-\sigma(z_{j,c})).
\end{equation}
Its gradient with respect to the input is
\begin{equation}
    \nabla_x \mathcal{L}_{\mathrm{CLS}}(j)
    =
    \big(\sigma(z_{j,y_{i^\star}})-1\big)\nabla_x z_{j,y_{i^\star}}
    +
    \sum_{c\neq y_{i^\star}}
    \sigma(z_{j,c}) \nabla_x z_{j,c}.
\end{equation}
Thus, maximising $\mathcal{L}_{\mathrm{CLS}}(j)$ tends to decrease the ground-truth logit while increasing the class-wise penalties associated with non-ground-truth logits. This makes classification-loss maximisation an indirect surrogate for the thresholded score events that define RMA and ODA.

By contrast, the score-level events associated with RMA and ODA are governed by extremal class scores:
\begin{equation}
    s^{\mathrm{oth}}_j=\max_{c\neq y_{i^\star}} s_{j,c},
    \qquad
    s^{\mathrm{max}}_j=\max_c s_{j,c}.
\end{equation}
Away from ties, the corresponding gradients depend only on the class attaining the relevant maximum:
\begin{equation}
    \nabla_x s^{\mathrm{oth}}_j = \nabla_x s_{j,c^\star},
    \quad
    c^\star \in \arg\max_{c\neq y_{i^\star}} s_{j,c},
    \qquad
    \nabla_x s^{\mathrm{max}}_j = \nabla_x s_{j,\tilde c^\star},
    \quad
    \tilde c^\star \in \arg\max_c s_{j,c}.
\end{equation}
At score ties, the same reasoning applies in the subgradient sense.

\paragraph{Proposition A3 (partial misalignment of class-loss maximisation).}
For additive multi-label classification heads, the update direction induced by maximising $\mathcal{L}_{\mathrm{CLS}}(j)$ generally need not coincide with a direction tailored to either the RMA score-threshold event or the ODA score-threshold event. Alignment occurs only under restrictive local conditions, for example when the relevant class-gradient contributions are collinear and have compatible signs.

\paragraph{Proof.}
We prove the claim by exhibiting local configurations in which the classification-loss ascent direction is not aligned with the natural score-event directions for RMA and ODA.

Consider the BCE illustration above. For a matched prediction $j$, the input-gradient of the classification loss is
\begin{equation}
    \nabla_x \mathcal{L}_{\mathrm{CLS}}(j)
    =
    \big(\sigma(z_{j,y})-1\big)\nabla_x z_{j,y}
    +
    \sum_{c\neq y}
    \sigma(z_{j,c})\nabla_x z_{j,c},
\end{equation}
where we write $y=y_{i^\star}$ for brevity. Thus, the classification-loss ascent direction combines the ground-truth logit gradient with a weighted sum of all non-ground-truth class-gradient contributions.

For RMA, the relevant score-threshold event is controlled by
\begin{equation}
    s^{\mathrm{oth}}_j=\max_{c\neq y}s_{j,c}.
\end{equation}
Away from ties, let $c^\star=\arg\max_{c\neq y}s_{j,c}$. A direction tailored to increasing the RMA margin is therefore aligned with
\begin{equation}
    \nabla_x s^{\mathrm{oth}}_j
    =
    \nabla_x s_{j,c^\star}.
\end{equation}
For sigmoid scores, $\nabla_x s_{j,c^\star}=\sigma'(z_{j,c^\star})\nabla_x z_{j,c^\star}$, so this direction is collinear with $\nabla_x z_{j,c^\star}$.

Now consider a local configuration with two non-ground-truth classes $c_1,c_2$, where $c_1=c^\star$ is the strongest competing class, and suppose
\begin{equation}
    \nabla_x z_{j,c_1}=e_1,\qquad
    \nabla_x z_{j,c_2}=e_2,\qquad
    \nabla_x z_{j,y}=0,
\end{equation}
with $e_1$ and $e_2$ linearly independent. Since $\sigma(z_{j,c_1})>0$ and $\sigma(z_{j,c_2})>0$, the classification-loss ascent direction contains the component
\begin{equation}
    \sigma(z_{j,c_1})e_1+\sigma(z_{j,c_2})e_2,
\end{equation}
which is not collinear with $e_1$. However, the RMA-tailored direction is collinear with $\nabla_x s^{\mathrm{oth}}_j$, and hence with $e_1$. Therefore, the classification-loss ascent direction need not coincide with the RMA-oriented direction.

For ODA, the relevant score-threshold event is controlled by
\begin{equation}
    s^{\mathrm{max}}_j=\max_c s_{j,c}.
\end{equation}
Away from ties, let $\tilde c^\star=\arg\max_c s_{j,c}$. A direction tailored to increasing the ODA margin $m^{\mathrm{ODA}}_j=\tau-s^{\mathrm{max}}_j$ is aligned with
\begin{equation}
    -\nabla_x s^{\mathrm{max}}_j
    =
    -\nabla_x s_{j,\tilde c^\star}.
\end{equation}
Suppose, for example, that the ground-truth class is currently the maximum, so $\tilde c^\star=y$. Let
\begin{equation}
    \nabla_x z_{j,y}=e_1,\qquad
    \nabla_x z_{j,c_1}=e_2,
\end{equation}
with $e_1$ and $e_2$ linearly independent. The ODA-tailored direction is collinear with $-e_1$. By contrast, the classification-loss ascent direction contains both
\begin{equation}
    \big(\sigma(z_{j,y})-1\big)e_1
    \quad\text{and}\quad
    \sigma(z_{j,c_1})e_2 .
\end{equation}
Since $\sigma(z_{j,c_1})>0$, this direction is not collinear with $-e_1$ unless the non-ground-truth contribution is absent or collinear with $e_1$. Thus, classification-loss maximisation is not generally aligned with the ODA-oriented direction either.

These examples show that alignment between classification-loss maximisation and the RMA or ODA score-threshold directions requires special local structure, such as collinearity of the relevant class-gradient contributions or negligible gradients from non-controlling classes. In general, classification-loss maximisation is therefore only an indirect surrogate for RMA- and ODA-oriented adversarial generation. \hfill $\square$

The same qualitative issue can arise for softmax-based class heads in set-based detectors: moving probability mass away from the correct class is not equivalent to directly controlling the thresholded extremal score events that define RMA and ODA.

\subsection{Role of the Concentration Penalty}
\label{app:theory_concentration}

The concentration term
\begin{equation}
    \mathcal{L}_{\mathrm{CON}}(j)
    =
    -\sum_{c\neq y_{i^\star}} \bar s_{j,c}\log(\bar s_{j,c}+\zeta)
\end{equation}
is designed to discourage the RMA branch from distributing non-target score mass diffusely across many incorrect classes.

When the total non-target score is positive, we define the normalised non-target scores as
\begin{equation}
    \bar s_{j,c}
    =
    \frac{s_{j,c}}
    {\sum_{c'\neq y_{i^\star}} s_{j,c'}},
    \qquad c\neq y_{i^\star}.
\end{equation}
Ignoring the small stabilising constant $\zeta$ for the moment, $\mathcal{L}_{\mathrm{CON}}(j)$ is the Shannon entropy of the normalised non-target score vector $\bar{\mathbf{s}}_{j,\neg y}$. Minimising this entropy encourages the non-target score mass to concentrate on a single incorrect class rather than spreading across multiple classes. This is consistent with an RMA-style misclassification event, where a dominant incorrect class should exceed the detection threshold. If all non-target scores are zero, we set $\mathcal{L}_{\mathrm{CON}}(j)=0$ by convention, since there is then no non-target mass to concentrate.

\paragraph{Proposition A4 (entropy minimisation encourages a dominant false class).}
Let $\bar{\mathbf{s}}_{j,\neg y} \in \Delta^{C-2}$ denote the normalised non-target score vector. Then
\begin{equation}
    0 \le
    -\sum_{c\neq y_{i^\star}} \bar s_{j,c}\log \bar s_{j,c}
    \le
    \log(C-1),
\end{equation}
with the lower bound attained if and only if all mass is concentrated on a single non-target class, and the upper bound attained if and only if the non-target distribution is uniform.

\paragraph{Proof.}
This is the standard extremal property of Shannon entropy on the probability simplex: entropy is minimised at the vertices of the simplex and maximised by the uniform distribution. \hfill $\square$

Thus, when the RMA branch is active, minimising $\mathcal{L}_{\mathrm{CON}}$ encourages the non-target score distribution to concentrate on a single incorrect class rather than spreading diffusely across multiple classes. This promotes an RMA-style misclassification pattern, where one incorrect class becomes dominant, which generic classification-loss maximisation does not explicitly enforce.

\subsection{Soft Minimisation as Entropy-Regularised Branch Selection}
\label{app:theory_softmin}

We now formalise the role of the soft gate in the soft minimisation objective. For a matched prediction $j$, define
\begin{equation}
    a_j = \mathcal{L}_{\mathrm{RMA}}(j), \qquad
    b_j = \mathcal{L}_{\mathrm{ODA}}(j), \qquad
    c_j = \mathcal{L}_{\mathrm{CON}}(j).
\end{equation}
The branch weights are computed as
\begin{equation}
    g_j^{\mathrm{RMA}},\,g_j^{\mathrm{ODA}}
    =
    \mathrm{softmax}(-a_j,-b_j).
\end{equation}
Thus, the gate assigns greater weight to the branch with the smaller loss, corresponding to the branch whose score-level condition is closer to being satisfied.

\paragraph{Proposition A5 (exact decomposition of the gated objective).}
For each matched prediction $j$, the gated RMA--ODA objective satisfies
\begin{equation}
    g_j^{\mathrm{RMA}} a_j + g_j^{\mathrm{ODA}} b_j
    =
    -\log\!\big(e^{-a_j}+e^{-b_j}\big)
    +
    H(g_j),
\end{equation}
where
\begin{equation}
    H(g_j)
    =
    - g_j^{\mathrm{RMA}}\log g_j^{\mathrm{RMA}}
    - g_j^{\mathrm{ODA}}\log g_j^{\mathrm{ODA}}
    \in [0,\log 2]
\end{equation}
is the entropy of the two-way gate. Consequently, since
\begin{equation}
    \mathcal{L}_{\mathrm{ADV}}(j)
    =
    g_j^{\mathrm{RMA}}(a_j+c_j)+g_j^{\mathrm{ODA}}b_j,
\end{equation}
we have
\begin{equation}
    \mathcal{L}_{\mathrm{ADV}}(j)
    =
    -\log\!\big(e^{-a_j}+e^{-b_j}\big)
    +
    H(g_j)
    +
    g_j^{\mathrm{RMA}} c_j .
\end{equation}

\paragraph{Proof.}
Let
\begin{equation}
    Z_j=e^{-a_j}+e^{-b_j}.
\end{equation}
By the definition of the softmax gate,
\begin{equation}
    g_j^{\mathrm{RMA}}=\frac{e^{-a_j}}{Z_j},
    \qquad
    g_j^{\mathrm{ODA}}=\frac{e^{-b_j}}{Z_j}.
\end{equation}
Therefore,
\begin{equation}
    -\log g_j^{\mathrm{RMA}}
    =
    a_j+\log Z_j,
    \qquad
    -\log g_j^{\mathrm{ODA}}
    =
    b_j+\log Z_j.
\end{equation}
Multiplying these identities by the corresponding gate weights and summing gives
\begin{align}
    H(g_j)
    &=
    -g_j^{\mathrm{RMA}}\log g_j^{\mathrm{RMA}}
    -
    g_j^{\mathrm{ODA}}\log g_j^{\mathrm{ODA}} \nonumber\\
    &=
    g_j^{\mathrm{RMA}}a_j
    +
    g_j^{\mathrm{ODA}}b_j
    +
    \log Z_j .
\end{align}
Rearranging yields
\begin{equation}
    g_j^{\mathrm{RMA}}a_j
    +
    g_j^{\mathrm{ODA}}b_j
    =
    -\log Z_j + H(g_j)
    =
    -\log\!\big(e^{-a_j}+e^{-b_j}\big)+H(g_j).
\end{equation}
Substituting this identity into
\begin{equation}
    \mathcal{L}_{\mathrm{ADV}}(j)
    =
    g_j^{\mathrm{RMA}}(a_j+c_j)+g_j^{\mathrm{ODA}}b_j
\end{equation}
gives the stated decomposition of $\mathcal{L}_{\mathrm{ADV}}(j)$. \hfill $\square$

\paragraph{Corollary A5.1 (soft lower envelope).}
For every matched prediction $j$,
\begin{equation}
    -\log\!\big(e^{-a_j}+e^{-b_j}\big)
    \le
    g_j^{\mathrm{RMA}} a_j + g_j^{\mathrm{ODA}} b_j
    \le
    -\log\!\big(e^{-a_j}+e^{-b_j}\big)+\log 2 .
\end{equation}
Moreover, as $|a_j-b_j|\to\infty$, the gate entropy satisfies $H(g_j)\to 0$, so the gated average approaches the log-sum-exp soft minimum. In the same limit, this soft minimum approaches $\min\{a_j,b_j\}$.

This corollary clarifies the role of the gate. The branch-selection component is not merely a heuristic average of the RMA and ODA losses; it is an entropy-regularised smooth selector that assigns greater weight to the branch closer to satisfaction. When one branch is clearly easier to satisfy, the gate becomes nearly deterministic. The additional term $g_j^{\mathrm{RMA}}c_j$ applies the concentration penalty in proportion to the activation of the RMA branch.

\subsection{Repaired Margin Induced by the Outer Defence Terms}
\label{app:theory_outer}

The outer loss augments the standard detection loss with
\begin{equation}
    \mathcal{L}_{\mathrm{REC}}(j)=\mathrm{softplus}(\tau-s^{\mathrm{gt}}_j),
    \qquad
    \mathcal{L}_{\mathrm{SUP}}(j)=\mathrm{softplus}(s^{\mathrm{oth}}_j-\tau).
\end{equation}
These terms target the score inequalities that counter ODA and RMA at the matched-prediction level: the recovery term encourages the ground-truth score to exceed the detection threshold, while the suppression term encourages all non-ground-truth scores to remain below it.

We define the repaired score margin as
\begin{equation}
    m_j(x')
    =
    \min\{s^{\mathrm{gt}}_j(x')-\tau,\; \tau-s^{\mathrm{oth}}_j(x')\}.
\end{equation}

\paragraph{Proposition A6 (positive repaired margin excludes both score-level failure modes).}
If $m_j(x')>0$ for some matched prediction $j\in \mathcal{J}_{i^\star}(x')$, then
\begin{equation}
    s^{\mathrm{gt}}_j(x')>\tau
    \qquad\text{and}\qquad
    s^{\mathrm{oth}}_j(x')<\tau.
\end{equation}
Therefore, at the pre-NMS score-filtering stage, this prediction survives score thresholding under the ground-truth class and has no threshold-crossing non-ground-truth class score.

\paragraph{Proof.}
The result follows directly from the definition
\[
    m_j(x')=\min\{s^{\mathrm{gt}}_j(x')-\tau,\; \tau-s^{\mathrm{oth}}_j(x')\}.
\]
If $m_j(x')>0$, then both arguments of the minimum are positive. Hence
\[
    s^{\mathrm{gt}}_j(x')-\tau>0
    \qquad\text{and}\qquad
    \tau-s^{\mathrm{oth}}_j(x')>0,
\]
which gives $s^{\mathrm{gt}}_j(x')>\tau$ and $s^{\mathrm{oth}}_j(x')<\tau$. \hfill $\square$

Proposition~A6 is intentionally modest but useful. It shows that the added defence terms encourage precisely the score margin that the generic detection loss does not explicitly enforce: the recovery term pushes the correct class above threshold, while the suppression term pushes competing classes below threshold.

Connecting this score-level statement to the final detector output requires a local stability assumption on localisation, matching, and post-processing.

\paragraph{Corollary A6.1 (local score stability under fixed-match/fixed-box conditions).}
Suppose that, in a neighbourhood $\mathcal{U}$ of $x'$, the following conditions hold for some matched prediction $j$:
\begin{enumerate}
    \item the prediction remains matched to the target object, and its box remains compatible with the IoU and NMS conditions used at inference; and
    \item the score functions $s^{\mathrm{gt}}_j(\cdot)$ and $s^{\mathrm{oth}}_j(\cdot)$ are $L$-Lipschitz on $\mathcal{U}$ with respect to the perturbation norm.
\end{enumerate}
If $m_j(x')=\gamma>0$, then for any additional perturbation $\eta$ supported inside the target box and satisfying $\|\eta\|<\gamma/L$, the repaired score inequalities are preserved:
\begin{equation}
    s^{\mathrm{gt}}_j(x'+\eta)>\tau,
    \qquad
    s^{\mathrm{oth}}_j(x'+\eta)<\tau.
\end{equation}

\paragraph{Proof.}
Since $m_j(x')=\gamma$, we have
\[
    s^{\mathrm{gt}}_j(x')\ge \tau+\gamma,
    \qquad
    s^{\mathrm{oth}}_j(x')\le \tau-\gamma.
\]
By Lipschitz continuity,
\begin{equation}
    s^{\mathrm{gt}}_j(x'+\eta)
    \ge
    s^{\mathrm{gt}}_j(x') - L\|\eta\|
    >
    \tau+\gamma-\gamma
    =
    \tau,
\end{equation}
and similarly,
\begin{equation}
    s^{\mathrm{oth}}_j(x'+\eta)
    \le
    s^{\mathrm{oth}}_j(x') + L\|\eta\|
    <
    \tau-\gamma+\gamma
    =
    \tau.
\end{equation}
Thus, both repaired score inequalities are preserved. \hfill $\square$

This corollary gives a local stability interpretation of the outer loss: larger repaired margins provide greater tolerance to additional perturbations at the score level, conditional on localisation, matching, and post-processing remaining stable.

\subsection{Expected-Informativeness Justification for Filtered Weighted Selection}
\label{app:theory_fws}

Filtered weighted selection (FWS) allocates the adversarial-generation budget to target objects that are already cleanly detected. This helps avoid cases where the inner loop primarily exploits pre-existing detection ambiguity rather than backdoor-related behaviour.

Let $\mathcal{I}_{\mathrm{clean}}$ denote the set of candidate objects that pass the clean-detection filter. For each $i\in\mathcal{I}_{\mathrm{clean}}$, let $c_i>0$ denote its clean confidence, and let $q_i\in[0,1]$ denote the probability that selecting object $i$ yields an informative adversarial example. Here, ``informative'' refers to cases where the selected object provides stable matching, unambiguous supervision, or a repair gradient more likely to target backdoor behaviour than ordinary detection failure.

The following argument justifies the confidence-weighting step conditional on the filtered candidate set. Under uniform selection over $\mathcal{I}_{\mathrm{clean}}$,
\begin{equation}
    \mathbb{E}_{\mathrm{unif}}[q]
    =
    \frac{1}{|\mathcal{I}_{\mathrm{clean}}|}
    \sum_{i\in\mathcal{I}_{\mathrm{clean}}} q_i .
\end{equation}
Under confidence-weighted selection with
\begin{equation}
    p_i
    =
    \frac{c_i}
    {\sum_{k\in\mathcal{I}_{\mathrm{clean}}} c_k},
\end{equation}
the expected informativeness becomes
\begin{equation}
    \mathbb{E}_{\mathrm{FWS}}[q]
    =
    \sum_{i\in\mathcal{I}_{\mathrm{clean}}}p_i q_i
    =
    \frac{\sum_{i\in\mathcal{I}_{\mathrm{clean}}} c_i q_i}
    {\sum_{i\in\mathcal{I}_{\mathrm{clean}}} c_i}.
\end{equation}

\paragraph{Proposition A7 (confidence weighting improves expected informativeness under positive correlation).}
If the covariance between clean confidence and informativeness is nonnegative over $\mathcal{I}_{\mathrm{clean}}$, then
\begin{equation}
    \mathbb{E}_{\mathrm{FWS}}[q]
    \ge
    \mathbb{E}_{\mathrm{unif}}[q].
\end{equation}

\paragraph{Proof.}
Let $\bar c$ and $\bar q$ denote the uniform averages of $\{c_i\}$ and $\{q_i\}$ over $\mathcal{I}_{\mathrm{clean}}$. Since $c_i>0$ for all $i\in\mathcal{I}_{\mathrm{clean}}$, we have $\bar c>0$. Then
\begin{equation}
    \mathbb{E}_{\mathrm{FWS}}[q]
    =
    \frac{\mathbb{E}_{\mathrm{unif}}[cq]}
    {\mathbb{E}_{\mathrm{unif}}[c]}
    =
    \frac{\mathrm{Cov}_{\mathrm{unif}}(c,q)+\bar c\,\bar q}
    {\bar c}
    =
    \bar q + \frac{\mathrm{Cov}_{\mathrm{unif}}(c,q)}{\bar c}.
\end{equation}
Therefore, if $\mathrm{Cov}_{\mathrm{unif}}(c,q)\ge 0$, then
\begin{equation}
    \mathbb{E}_{\mathrm{FWS}}[q]
    \ge
    \bar q
    =
    \mathbb{E}_{\mathrm{unif}}[q].
\end{equation}
\hfill $\square$

This proposition captures the role of the weighting step within the filtered candidate pool. If objects with higher clean confidence are more likely to yield stable and informative target-object supervision, then confidence-weighted selection improves the expected quality of the selected inner-loop target. Relative to unfiltered random selection, FWS also excludes objects that fail the clean-detection filter by construction. Proposition~A7 isolates the additional effect of confidence weighting among the retained candidates.

\subsection{Summary of Analytical Findings}
\label{app:theory_summary}

The analysis above supports four main points.

First, the proposed inner and outer losses are monotone-aligned with the score-threshold margins that describe object-level backdoor attack and repair events. Second, restricting optimisation to $\mathcal{J}_{i^\star}(x'(\delta))$ reduces gradient dilution from unrelated detections under mild alignment assumptions. Third, the soft branch minimisation objective can be interpreted as an entropy-regularised smooth selector between the RMA and ODA branches, while the concentration term encourages a dominant incorrect class when the RMA branch is active. Fourth, the outer defence terms induce a repaired score margin that excludes both score-level failure modes and provides local score stability once such a margin is achieved.

These results should be interpreted as a mechanism-level justification rather than a global robustness guarantee for modern object detectors. Final detection outcomes also depend on localisation, matching, and post-processing, which are architecture-specific and highly nonlinear. The role of this analysis is therefore to justify the design of the proposed method and to complement, rather than replace, empirical evaluation.

\section{Hyperparameter Selection}\label{sec:appendix-selection-hyper}
In the following section we provide experiments that show how sensitive the proposed methods are to variations in the chosen hyperparameters. For these experiments, we use SBM-FWS. In the main text, as well as the additional experiments provided in the subsequent sections~\ref{sec:appendix-additional-ablations} and \ref{sec:appendix-additional-results}, the hyperparameters used for training are defined in Table~\ref{tab:hyperparameters}. In all cases, $\tau$ is set to the confidence threshold used by each model.

\begin{table}[h]
\centering
\caption{Model Configuration Summary.}
\label{tab:hyperparameters}
\resizebox{\textwidth}{!}{%
\begin{tabular}{lcccc|ccc|ccccc}
\hline
 & \multicolumn{4}{c|}{\textbf{Hyperparameters}} & \multicolumn{3}{c|}{\textbf{Design Choices}} & \multicolumn{5}{c}{\textbf{Training Parameters}} \\
\textbf{Model}
  & $\lambda$ & $\beta$ & $\epsilon$ & IoU & Train BB & $\theta_{\text{UP}}$ & Match.\ Strat.
  & LR & Epochs & Batch Size & Adv.\ Steps & Step Size \\
\midrule
YOLOv5 & 0.01 & 1.0 & 0.0314 & 0.60 & No & No  & Yes & 0.000025 & 30 & 16 & 30 & 0.0078 \\
DINO   & 0.1  & 1.0 & 0.0314 & 0.60 & No & Yes & No  & 0.000091 & 30 & 16 & 30 & 0.0078 \\
FCOS   & 0.1  & 0.5 & 0.0314 & 0.60 & No & No  & Yes & 0.001    & 30 & 16 & 30 & 0.0078 \\
\hline
\end{tabular}%
}
\end{table}

\subsection{Lambda ($\lambda$)}
\begin{figure}[h]
    \centering
    \includegraphics[width=\linewidth]{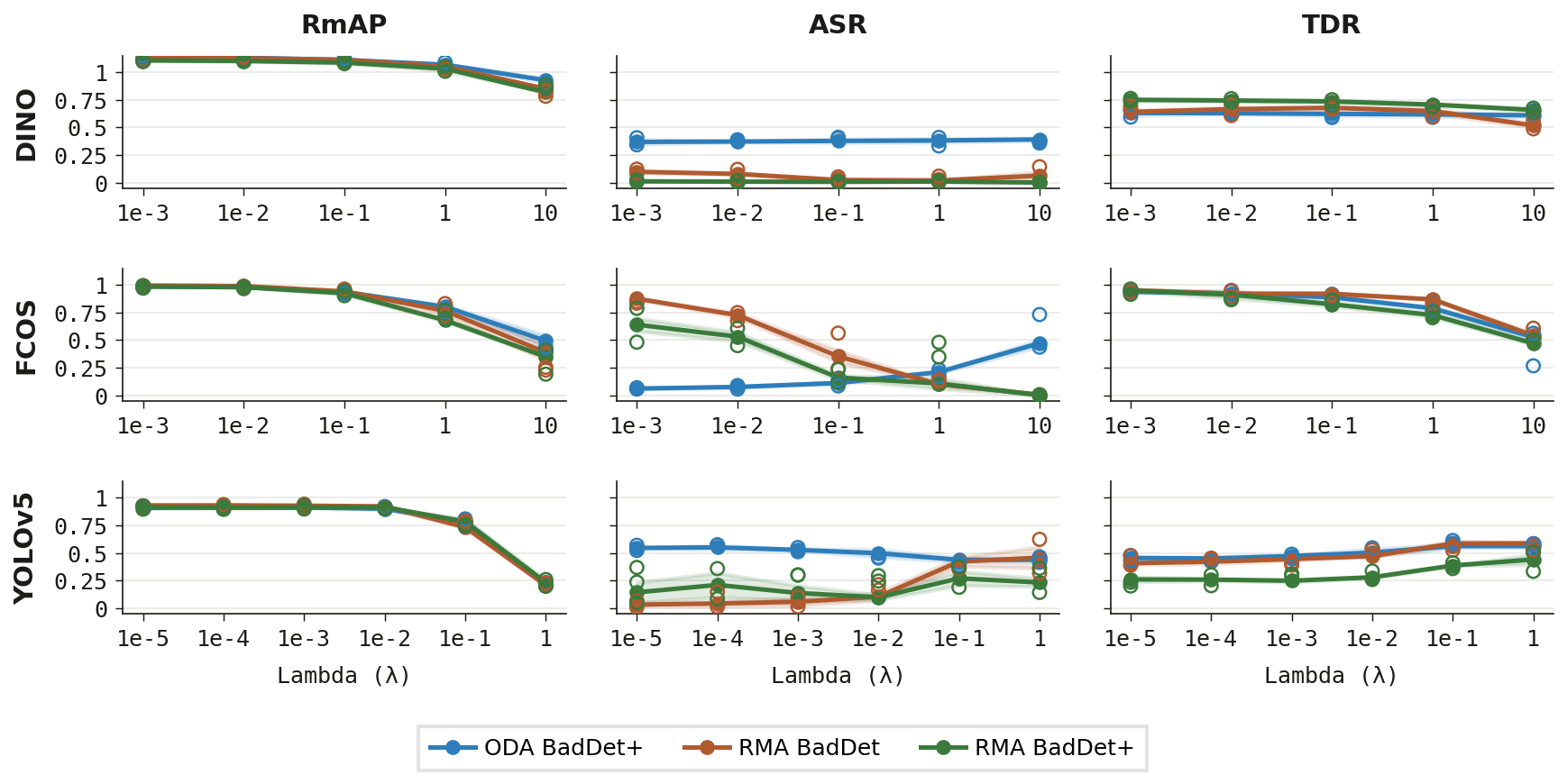}
    \caption{Performance of each model when $\lambda$ is varied. High RmAP and TDR are better, while low ASR is preferred.}
    \label{fig:lambda}
\end{figure}

In Section~\ref{sec:advft}, we proposed adding the term $\mathcal{L}_{\text{REC}}$ and $\mathcal{L}_{\text{SUP}}$ when fine-tuning the model on adversarial examples. To integrate this penalty with the normal detector losses $\mathcal{L}_{\text{OD}}$, $\lambda$ is used to balance its contribution. For each model architecture, we evaluate the impact that $\lambda$ has on performance across each model architecture and for each attack objective. We show the results in Fig.~\ref{fig:lambda}. In general, we find that the value of $\lambda$ has smooth performance changes as its value increases and decreases. Moreover, we find that as $\lambda$ increases (i.e., the strength of the proposed defence penalty increases), the ASR-RmAP is greater than at lower value of $\lambda$, with this specifically being true for FCOS. For FCOS, BadDet+ ODA, the opposite is true, as more RmAP is traded off, ASR increases. This is largely due to large reductions in main task performances, influencing the model's ability to detect objects. For DINO and FCOS, $\lambda = 0.1$ is optimal, while for YOLOv5, $\lambda=0.01$ is optimal.

\subsection{Beta ($\beta$)}

\begin{figure}[h]
    \centering
    \includegraphics[width=\linewidth]{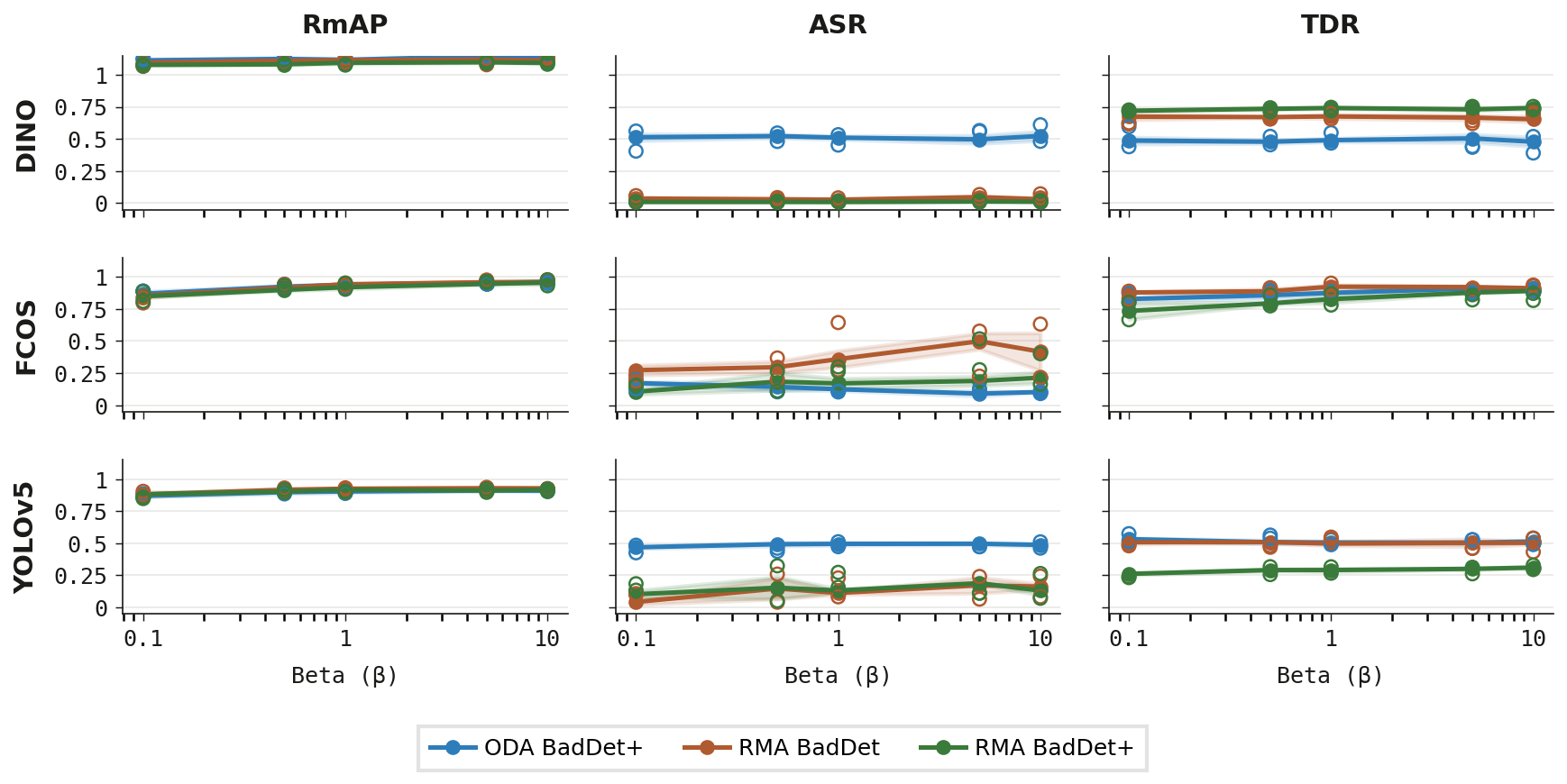}
    \caption{Performance of each model when $\beta$ is varied. High RmAP and TDR are better, while low ASR is preferred.}
    \label{fig:beta}
\end{figure}

In Section~\ref{sec:smin}, we use a $\mathrm{softplus}(\cdot)$ as part of the proposed term $\mathcal{L}_{\text{REC}}$ and $\mathcal{L}_{\text{SUP}}$, as well as in Section~\ref{sec:advft} as part of $\mathcal{L}_{\text{ODA}}$ and $\mathcal{L}_{\text{RMA}}$. As a result, $\beta$ is used in each definition to control the shape of the penalty in each case. For each model architecture, we evaluate the impact of $\beta$ on performance for each attack objective. We show the results in Fig.~\ref{fig:beta}. In general, we find that the value of $\beta$ has little impact on the performance of each model architecture. Similar to $\lambda$, we observe an ASR-RmAP trade-off when $\beta$ decreases and the softplus penalty becomes less piecewise-linear near the threshold $\tau$. For DINO and YOLOv5, we use the default value of $\beta=1$ and use $\beta=0.5$ in the case of FCOS.

\subsection{Epsilon ($\epsilon$)}

\begin{figure}[h]
    \centering
    \includegraphics[width=\linewidth]{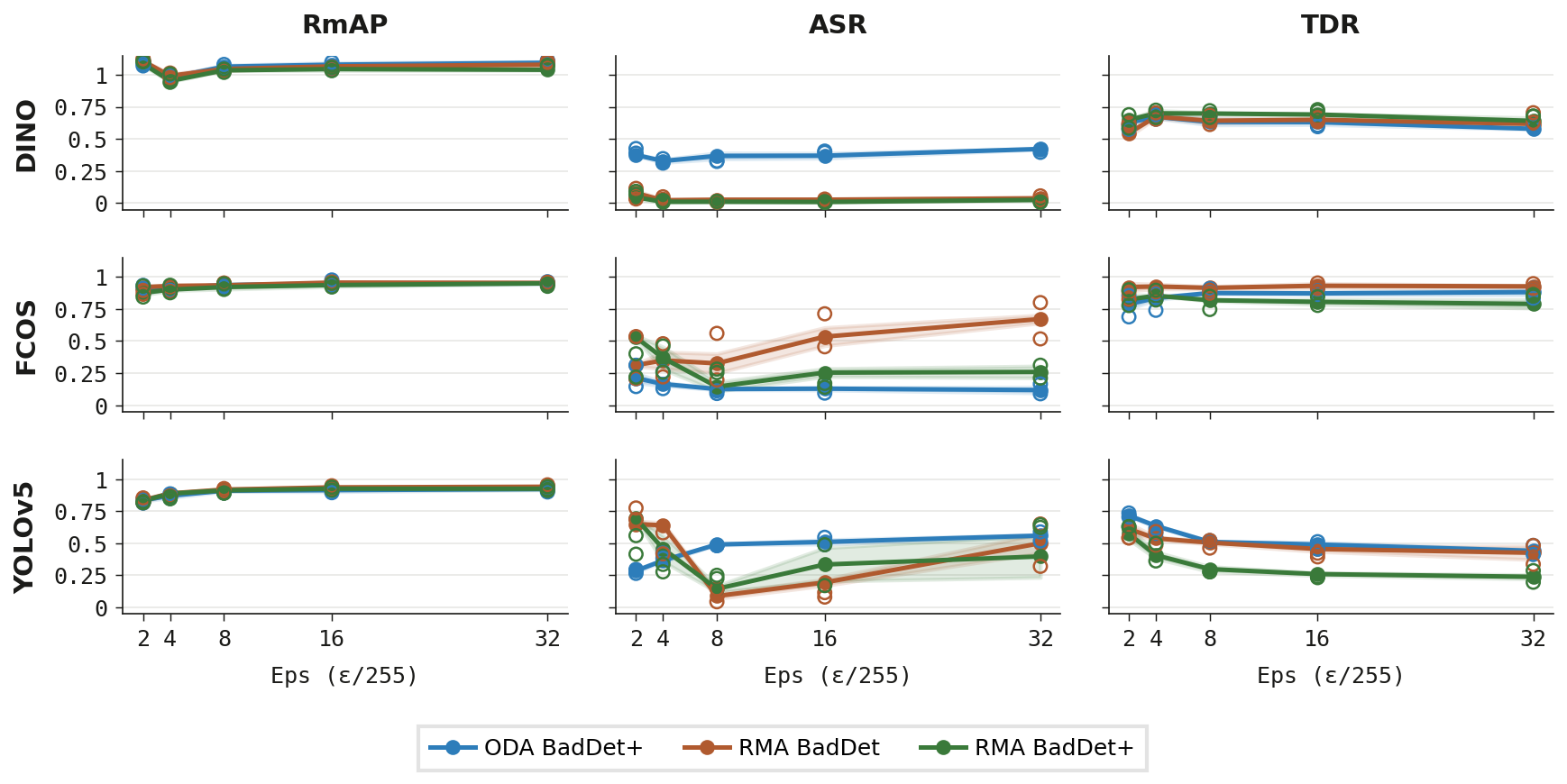}
    \caption{Performance of each model when $\epsilon$ is varied. High RmAP and TDR are better, while low ASR is preferred.}
    \label{fig:eps}
\end{figure}

As part of both the proposed adversarial objectives proposed in Section~\ref{sec:advgen}, an upper bound is placed on the perturbation $\delta$. As a result, the value of $\epsilon$ has a critical impact on the possible values of $\delta$ that are possible and thus the subsequent effectiveness of subsequent minimisation. We show the results in Fig.~\ref{fig:eps}. For DINO, we find that the value of $\epsilon$ is stable under both small and large values. In contrast, the FCOS and YOLO results are much more varied. In general, smaller values of $\epsilon$ provide limited ASR reduction. Performance improves as $\epsilon$ increases to $8$, but degrades again when $\epsilon$ is increased further to $32$. As a result, we find that $\epsilon=8$ offers the best trade-off for all three architectures, and that smaller and larger perturbations either do not allow for solutions of $\delta$ that facilitate classification errors that align with the attack, or expand the set of possible solutions to be so large that the number of useful solutions (i.e., solutions that provide information about the attack) is diluted by adversarial examples that are unrelated to the attack.

\subsection{IoU Value}

\begin{figure}[h]
    \centering
    \includegraphics[width=\linewidth]{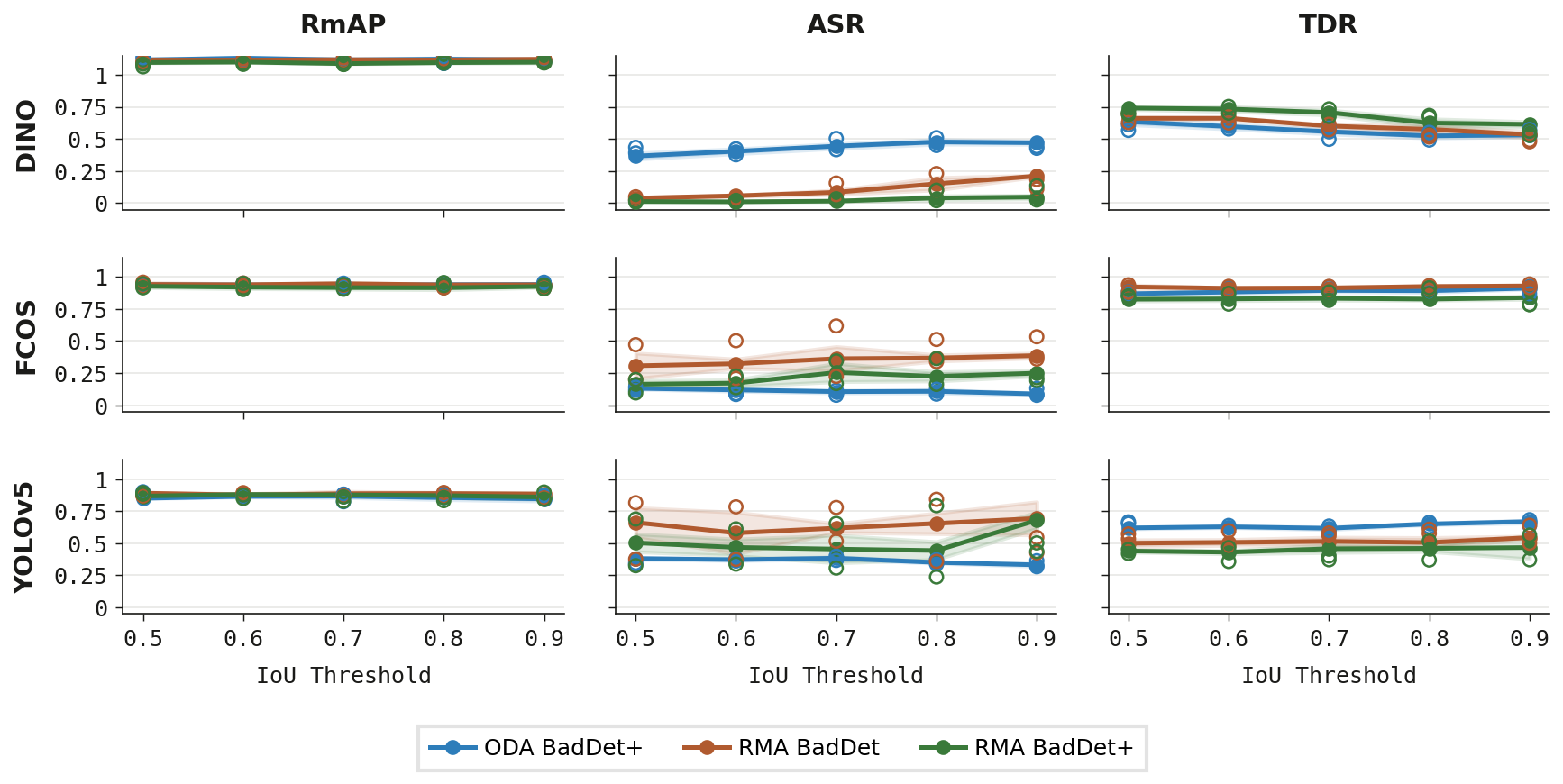}
    \caption{Performance of each model when the IoU threshold used by FWS is varied. High RmAP and TDR are better, while low ASR is preferred.}
    \label{fig:iou}
\end{figure}

In Section~\ref{sec:overall_opt}, we propose the Filtered weighted selection (FWS) strategy. As part of this strategy, an IoU threshold is used to select objects that have positive predictions. As a result, this threshold has an impact on how strict this selection is, and subsequently has the potential to impact performance. We show the results in Fig.~\ref{fig:iou}. In general, we find that IoU values above 0.5, the minimum value used by measures like mAP, cause small increases in ASR performance across most model architectures. While TDR shows slight increase for FCOS and YOLOv5, DINO shows some decrease in TDR as the IoU threshold increases. As a result, we chose a value of 0.6 in all experiments. 

\section{Additional Ablations} \label{sec:appendix-additional-ablations}
In the subsections below, we provide results that justify particular design choices made as part of our implementation. As part of our evaluation, we compare the performance of SBM-FWS when factors like the backbone are fine-tuned, the updated parameters ($\theta_{\text{UP}}$) are used to generate adversarial examples, and the DINO matching strategy is replaced with a strategy more similar to FCOS and YOLOv5. Moreover, we also show the performance difference between using the full loss (FLM) instead of just the classification loss in CLM, and demonstrate that FLM is equivalent to CLM in most cases.

\subsection{Backbone Training}

\begin{figure}[t]
    \centering
    % Left subfigure aligned at the top
    \begin{subfigure}[t]{0.35\textwidth}
        \centering
        \includegraphics[width=\linewidth]{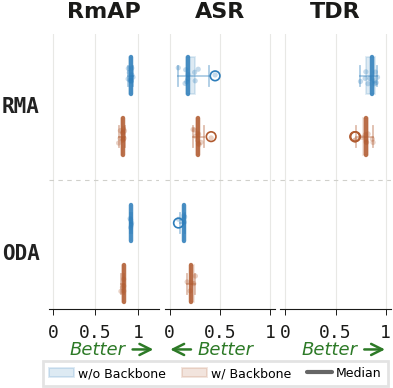}
        \caption{FCOS}
        \label{fig:ptsd_oda}
    \end{subfigure}
    \hfill 
    \begin{subfigure}[t]{0.31\textwidth}
        \centering
        \includegraphics[width=\linewidth]{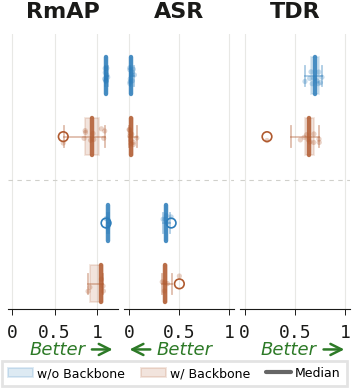}
        \caption{DINO}
    \end{subfigure}
    \hfill 
    \begin{subfigure}[t]{0.31\textwidth}
        \centering
        \includegraphics[width=\linewidth]{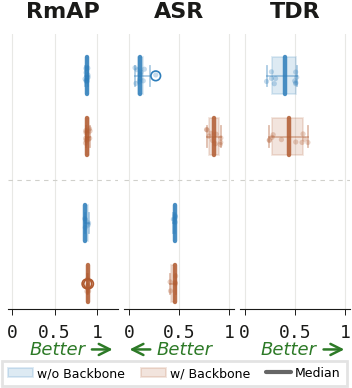}
        \caption{YOLOv5}
    \end{subfigure}
\caption{Performance of each model when the backbone is included and excluded from fine-tuning.}
\label{fig:backbone}
\end{figure}

The evaluated architectures employ a convolutional neural network (CNN) backbone, pretrained on large-scale image classification tasks, for robust feature extraction. To address scale invariance, these features are processed through a Feature Pyramid Network (FPN), which supplies multi-scale representations to a task-specific detection head. As a result, when employing the proposed adversarial fine-tuning, we considered if it is beneficial to fine-tune the entire model or just the detection head. In Fig.~\ref{fig:backbone} we show the performance of SBM-FWS under each paradigm. In the case of YOLOv5, substantial ASR benefits are gained when the backbone is not fine-tuned, particularly in both the RMA BadDet and BadDet+ cases. For FCOS, some modest RmAP, ASR, and TDR gains are made in all three cases. For DINO, ASR and TDR performance is similar in both cases; however, some increased RmAP and TDR variance is observed. As a result, we do not fine-tune the backbone for any of the considered model architectures.

\subsection{Adversarial Generation}

When solving the inner loop problem defined in Section~\ref{sec:overall_opt}, we must choose whether to generate adversarial examples at each step using the original model parameters ($\theta_{\text{ORG}}$) or the updated parameters ($\theta_{\text{UP}}$). If $\theta_{\text{ORG}}$ is used, perturbations ($\delta$) are sampled based on their effect on the original model, and $\theta_{\text{UP}}$ is subsequently updated to minimize the impact of $\delta$. Conversely, if $\theta_{\text{UP}}$ is used for both steps, $\delta$ is crafted to cause misclassification or disappearance against the current $\theta_{\text{UP}}$. The parameters $\theta_{\text{UP}}$ are then updated again to mitigate this new adversarial effect.

\begin{figure}[t]
    \centering
    % Left subfigure aligned at the top
    \begin{subfigure}[t]{0.35\textwidth}
        \centering
        \includegraphics[width=\linewidth]{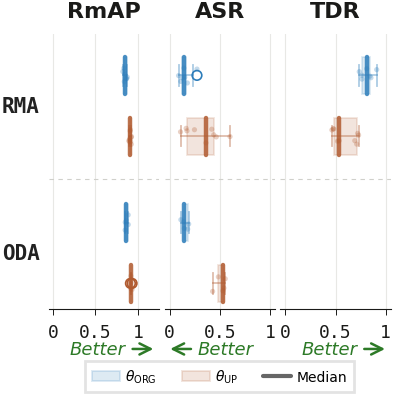}
        \caption{FCOS}
        \label{fig:ptsd_oda}
    \end{subfigure}
    \hfill 
    \begin{subfigure}[t]{0.31\textwidth}
        \centering
        \includegraphics[width=\linewidth]{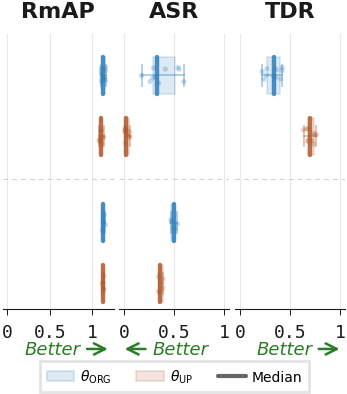}
        \caption{DINO}
    \end{subfigure}
    \hfill 
    \begin{subfigure}[t]{0.31\textwidth}
        \centering
        \includegraphics[width=\linewidth]{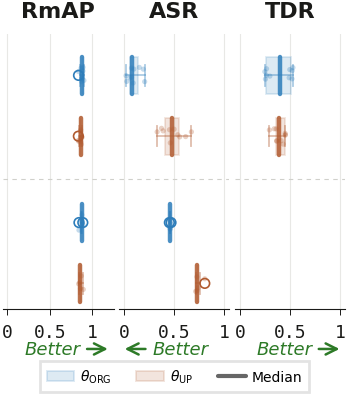}
        \caption{YOLOv5}
    \end{subfigure}
\caption{Performance of each model when $\theta_{\text{ORG}}$ or $\theta_{\text{UP}}$ is used to generate the adversarial examples.}
\label{fig:adv_gen}
\end{figure}

In Fig.~\ref{fig:adv_gen} we show the performance of SBM-FWS under each paradigm. For FCOS and YOLOv5, we find that using $\theta_{\text{ORG}}$ provides significant ASR and TDR performance improvements, while for DINO, the opposite is true. In the case of FCOS and YOLOv5, $\theta_{\text{ORG}}$ provides a biased estimate of the true trigger distribution, and adversarial fine-tuning succeeds. Moreover, when $\theta_{\text{UP}}$ is used instead, the initial mitigation of the backdoor is achieved when $\theta_{\text{UP}} \approx \theta_{\text{ORG}}$; however, after multiple iterations, this effect is lost due to catastrophic forgetting. That is, $\theta_{\text{UP}}$ introduces a moving target, causing adversarial examples to drift away from their backdoor objective over multiple iterations. For DINO, the opposite is true; using $\theta_{\text{UP}}$ is optimal. This is potentially due to the transformer detection head of DINO being more sensitive to adversarial examples, and thus the moving target introduced by $\theta_{\text{UP}}$ is helpful, or that it is less susceptible to catastrophic forgetting compared to FCOS and YOLOv5.

\subsection{DINO Matching Strategy}
\begin{figure}[t]
    \centering
    \begin{subfigure}[t]{0.62\textwidth}
        \centering
        \includegraphics[width=\linewidth]{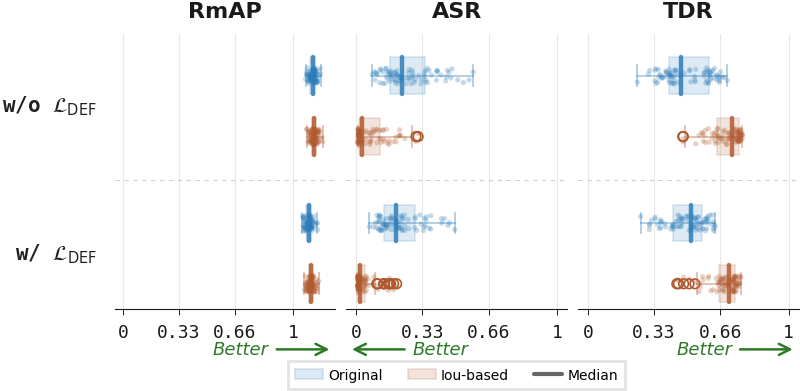}
        \caption{RMA}
    \end{subfigure}
    \hfill
    \begin{subfigure}[t]{0.35\textwidth}
        \centering
        \includegraphics[width=\linewidth]{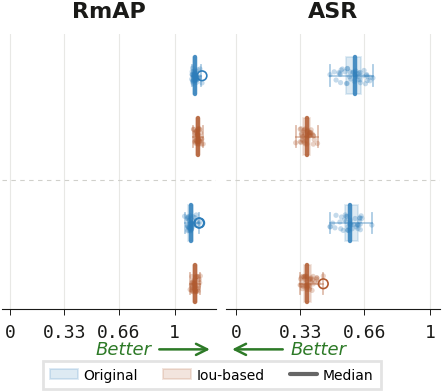}
        \caption{ODA}
    \end{subfigure}
\caption{Performance of DINO when its original and IoU-based matching strategy.}
\label{fig:dino-compare}
\end{figure}

In Fig.~\ref{fig:dino-compare}, we compare the performance when DINO's original matching strategy and a simple IoU-based matching strategy is used when solving the inner objective. Note, the IoU-based matching strategy matches all predictions with IoU>0.5 with the targeted object. For both ODA and RMA, as well as when the defence term is an is not included, using the IoU-based matching strategy significantly improves ASR and TDR performance, while also providing some RmAP gains. 

This improvement can be attributed to a fundamental incompatibility between DINO's Hungarian matching strategy and adversarial objectives that manipulate classification outputs. Hungarian matching jointly considers both localisation and classification scores when assigning predictions to ground-truth objects, meaning that as the adversarial objective alters a prediction's classification, the matched set of predictions can shift between optimisation steps. This creates a moving-target problem: the gradient updates intended to suppress or redirect a particular prediction may instead act on a different set of predictions in the next iteration, destabilising the attack. An IoU-based strategy avoids this by assigning matches purely on geometric overlap, decoupling the matching process from the classification output being manipulated. This is consistent with the design of FCOS and YOLOv5, whose assignment strategies are similarly geometry-driven and classification-agnostic, providing a more stable set of matched predictions throughout optimisation.

\subsection{Comparison of FLM and CLM} \label{sec:appendix-flm-vs-clm}
\begin{figure}[t]
    \centering
    % Left subfigure aligned at the top
    \begin{subfigure}[t]{0.3\textwidth}
        \centering
        \includegraphics[width=\linewidth]{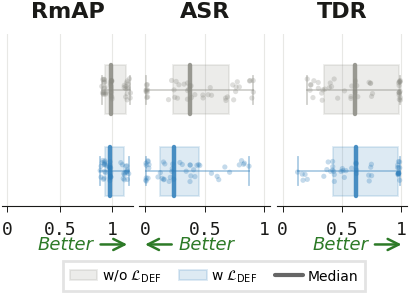}
        \caption{CLM}
        \label{fig:ptsd_oda}
    \end{subfigure}
    \hfill 
    \begin{subfigure}[t]{0.3\textwidth}
        \centering
        \includegraphics[width=\linewidth]{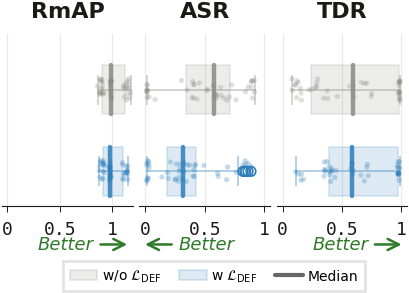}
        \caption{FLM}
    \end{subfigure}
    \hfill 
    \begin{subfigure}[t]{0.3\textwidth}
        \centering
        \includegraphics[width=\linewidth]{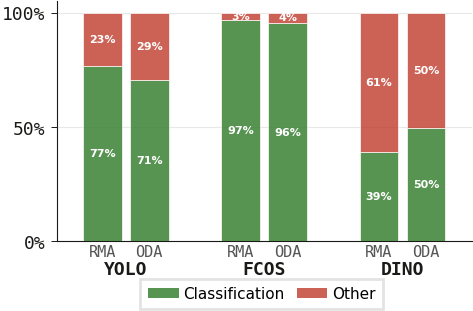}
        \caption{FLM}
    \end{subfigure}
\caption{Performance of (a) CLM and (b) FLM. (c) Shows the average contribution of the classification term to FLM relative to other terms.}
\label{fig:flm-clm-compare}
\end{figure}

In Figure~\ref{fig:flm-clm-compare} (a) and (b), we compare the performance of Classification Loss Maximization (CLM) against the alternative Full Loss Maximization (FLM) across three model architectures. For RMA attacks, constraining the maximization to the classification loss (CLM) yields a measurable performance improvement over maximizing the cumulative loss (FLM). For ODA, however, the performance gains are marginal.

Figure~\ref{fig:flm-clm-compare}(c) summarizes the mean loss deltas for FLM by model and attack objective. For FCOS and YOLOv5, the classification term dominates the cumulative loss during FLM, rendering FLM functionally similar to CLM. In contrast, DINO exhibits a more even distribution between classification and other loss terms. This discrepancy is likely rooted in model architecture: FCOS and YOLOv5 utilize distinct branches for classification and localization in the detection head, whereas DINO shares attention layers across these tasks.

Ultimately, filtering the maximization objective to classification terms only is a practical and effective design choice. For YOLO and FCOS, this filtering further concentrates gradients on the already dominant classification components. For DINO, it assists the perturbation $\delta$ in specifically targeting classification features. Critically, for both RMA and ODA, classification disruption is the primary objective, as unintended interference with localization terms is undesirable.

% \begin{table}[h]
% \centering
% \caption{Mean loss deltas ($\Delta$) for FLM across 64 batches (batch size 8). Delta represents the mean increase in loss terms following the adversarial perturbation.}
% \label{tab:loss_delta_summary}
% \begin{tabular}{lcccc}
% \toprule
% \textbf{Model} & \multicolumn{2}{c}{\textbf{RMA}} & \multicolumn{2}{c}{\textbf{ODA}} \\
% \cmidrule(lr){2-3} \cmidrule(lr){4-5}
%  & \textbf{Classification} & \textbf{Other} & \textbf{Classification} & \textbf{Other} \\
% \midrule
% YOLO & 1.441 & 0.436 & 1.419 & 0.593 \\
% FCOS & 16.354 & 0.558 & 13.767 & 0.634 \\
% DINO & 0.064 & 0.099 & 0.1 & 0.102 \\
% \bottomrule
% \end{tabular}
% \end{table}

\section{Additional Results} \label{sec:appendix-additional-results}

\subsection{COCO Evaluation}\label{sec:appendix-coco}
\begin{figure}
\centering
\includegraphics[width=0.6\textwidth]{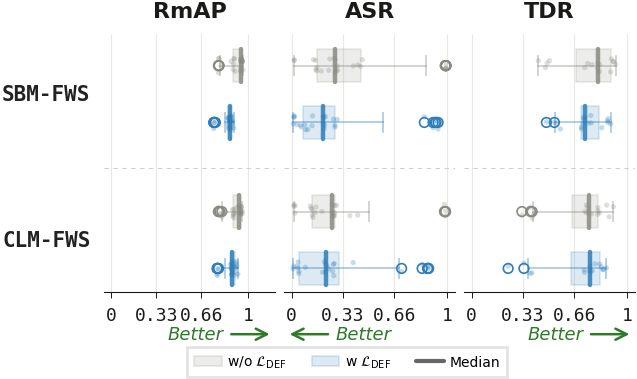}
\caption{Performance of CLM-FWS and SBM-FWS on COCO for RMA and ODA.}
\label{fig:coco}
\end{figure}

Fig.~\ref{fig:coco} shows the performance of CLM-FWS and SBM-FWS during a COCO evaluation, utilizing BadDet+ RMA and ODA attacks on the FCOS and DINO models. Due to the large number of classes present in COCO and the size of its original training dataset, a subset of 1000 images was used. In general, we find that both CLM-FWS and SBM-FWS are still effective, with median ASR performance remaining below 0.33 in both instances. Compared to the Pascal VOC results presented in Section~\ref{sec:results}, the result distributions are roughly similar. The COCO results, however, show slightly higher median ASR performance and longer variance in the lower tail when the defence term is applied.

Comparing the results with and without the defense term, we find that its inclusion is beneficial. For SBM-FWS, both the median ASR and its variance decrease when the defense term is included. For CLM-FWS, however, while the median performance marginally improves with the defense term, removing it yields a stronger lower-tail performance.

For both CLM-FWS and SBM-FWS, this evaluation demonstrates that the methods generalise to larger-scale object detection datasets. However, it also reveals that in cases where the models' baseline mAP performance is lower, as is the case for both DINO and FCOS compared to Pascal VOC and MTSD, mitigation via adversarial fine-tuning becomes more challenging. Specifically, it becomes more difficult to find high-quality adversarial examples that exploit backdoor behaviour, rather than exploiting a fragile detector that already produces brittle predictions on benign images.

\subsection{PTSD Evaluation}\label{sec:appendix-ptsd}
\begin{figure}[t]
    \centering
    % Left subfigure aligned at the top
    \begin{subfigure}[t]{0.45\textwidth}
        \centering
        \includegraphics[width=\linewidth]{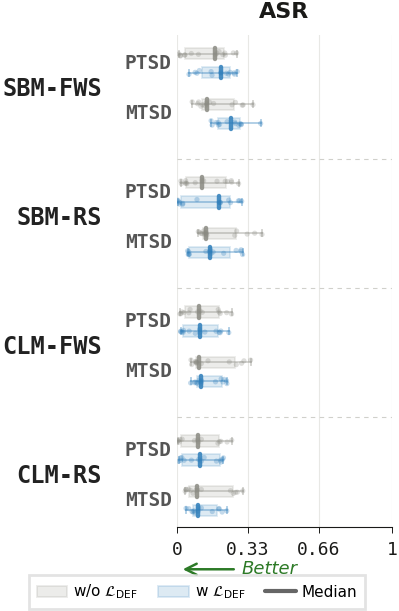}
        \caption{ODA}
        \label{fig:ptsd_oda}
    \end{subfigure}
    \hfill % Adds spacing between the two images
    % Right subfigure also aligned at the top
    \begin{subfigure}[t]{0.5\textwidth}
        \centering
        \includegraphics[width=\linewidth]{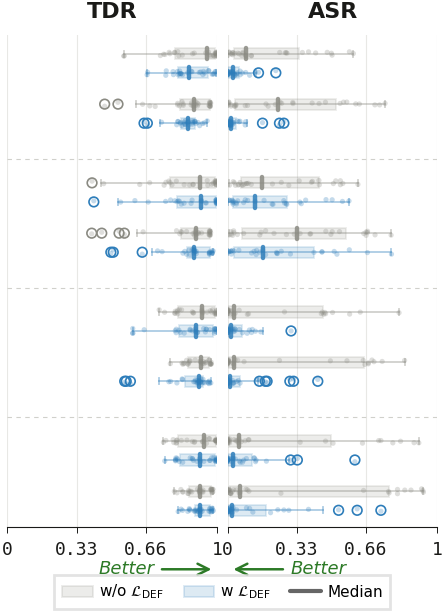}
        \caption{RMA}
        \label{fig:ptsd-rma}
    \end{subfigure}
\caption{Performance of each method on the PTSD and MTSD dataset.}
\label{fig:ptsd_overall}
\end{figure}

In Fig.~\ref{fig:ptsd_overall} we show the performance of each method on MTSD and PTSD. Note, PTSD is a collection of images containing physical triggers placed on real-world objects, and is used to validate whether a synthetic-to-real-world performance gap is present. In general, all methods show strong transferability between the two datasets, suggesting that MTSD serves as a reliable proxy for real-world trigger behaviour. For ASR, PTSD performance is typically the same or stronger than MTSD, indicating that physically realised triggers are at least as effective at inducing misclassification as their synthetic counterparts. For TDR, while median performance between MTSD and PTSD is typically similar, PTSD exhibits increased long-tail variance. This likely reflects real-world factors such as variation in lighting conditions, viewing angle, and print fidelity, which can introduce inconsistency in whether the defender successfully recovers correct classification, particularly in the case of RMA.

\begin{figure}[t]
    \centering
    \begin{subfigure}[t]{0.53\textwidth}
        \centering
        \includegraphics[width=\linewidth]{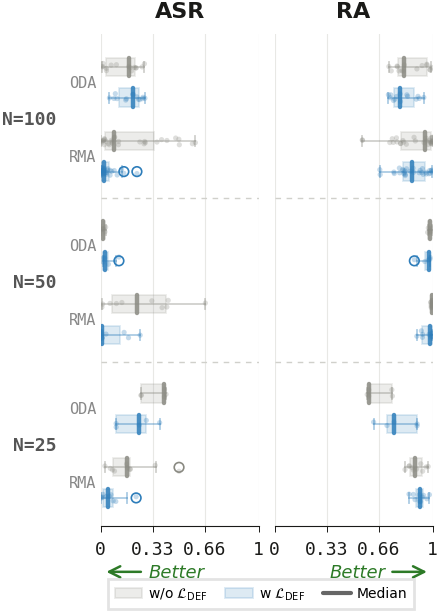}
        \caption{SBM-FWS}
    \end{subfigure}
    \hfill
    \begin{subfigure}[t]{0.43\textwidth}
        \centering
        \includegraphics[width=\linewidth]{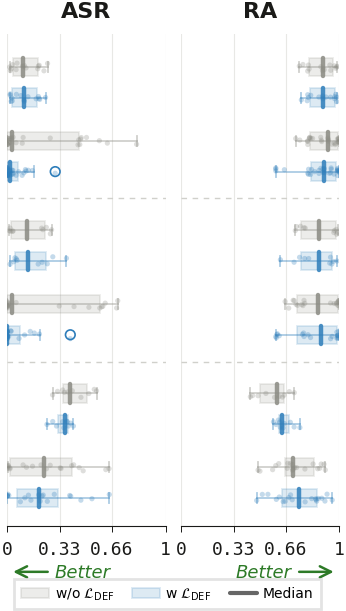}
        \caption{CLM-FWS}
    \end{subfigure}
\caption{Performance of CLM-FWS and SBM-FWS on the PTSD when data availability is reduced.}
\label{fig:ptsd_num_samples}
\end{figure}

\subsection{Computational Efficiency and Timing Analysis}

In Table~\ref{tab:timing}, we report the empirical computational cost breakdown for a single batch across our three evaluated architectures: FCOS, DINO, and YOLOv5. All experiments were conducted on a single H100 GPU to ensure consistency in timing measurements. This analysis serves to quantify the overhead introduced by the adversarial generation process and our proposed defense mechanism.

As expected in adversarial training and attack frameworks, the Adversarial Generation ($\delta$ optimization) consumes the vast majority of the computational budget, accounting for over 85--99\% of the total execution time per batch. Conversely, the inclusion of our proposed defense loss ($\mathcal{L}_{\mathrm{DEF}}$) adds negligible overhead (typically $<0.2\%$), demonstrating that the robustness gains reported in Section~\ref{sec:results} are essentially ``free'' in terms of training latency once the adversarial example is produced.

\textbf{Comparative Efficiency of FWS vs. RS:} A key observation is that FWS consistently yields lower total execution times compared to RS, despite FWS requiring an initial inference pass and a filtering heuristic. This efficiency stems from the quality of the targeted region in the inner loop optimization. By isolating a confidently predicted box and filtering out boxes containing false positives, the subsequent PGD steps benefit from speedup, particularly during the matching process used to isolate the predictions matched to the selected target object. While FWS incurs an initial ``filtering cost'' to identify the subset of predictions matching the targeted box, the resulting reduction during each PGD step means these savings compound, leading to a lower total time per batch compared to the RS.

\textbf{Architectural Variance:} We note a significant variance in absolute execution time between models. DINO, being a Transformer-based detector, exhibits the highest total cost (up to 6.15s per batch). We observe that DINO spends a lower relative percentage of time on adversarial generation compared to the CNN-based FCOS and YOLOv5. This is attributed to the inherent computational overhead of the self-attention mechanisms. Despite these differences in base architecture, the relative efficiency of FWS over RS remains consistent across all models.

\begin{table}[ht]
\centering
\caption{Computational cost breakdown per batch (averaged across 64 batches of size 8) for each defense variant across models. Adv Gen corresponds to the cost of performing the inner loop PGD steps. Def. Loss is the cost of calculating $\mathcal{L}_{\mathrm{DEF}}$. Total is the execution time in seconds.}
\label{tab:timing}
\resizebox{\textwidth}{!}{%
\begin{tabular}{l ccc ccc ccc}
\toprule
& \multicolumn{3}{c}{\textbf{FCOS}} & \multicolumn{3}{c}{\textbf{DINO}} & \multicolumn{3}{c}{\textbf{YOLOv5}} \\
\cmidrule(lr){2-4} \cmidrule(lr){5-7} \cmidrule(lr){8-10}
\textbf{Method} & Adv Gen & Def. Loss & Total & Adv Gen & Def. Loss & Total & Adv Gen & Def. Loss & Total \\
\midrule
CLM-RS   & 99.46\% & $<$0.01\% & 2.40 & 91.71\% & $<$0.01\% & 6.15 & 94.86\% & 0.12\% & 2.14 \\
CLM-FWS  & 98.68\% & $<$0.01\% & 2.27 & 85.89\% & $<$0.01\% & 4.11 & 89.63\% & 0.16\% & 1.35 \\
SBM-RS   & 96.38\% & $<$0.01\% & 2.76 & 91.97\% & $<$0.01\% & 6.10 & 94.41\% & $<$0.01\% & 2.02 \\
SBM-FWS  & 93.48\% & $<$0.01\% & 2.30 & 92.46\% & $<$0.01\% & 4.91 & 88.38\% & 0.11\% & 1.98 \\
\bottomrule
\end{tabular}}
\end{table}

\subsection{Additional Trigger Types} \label{appendix:triggers}
\begin{figure}[t]
    \centering
    \begin{subfigure}[t]{0.525\textwidth}
        \centering
        \includegraphics[width=\linewidth]{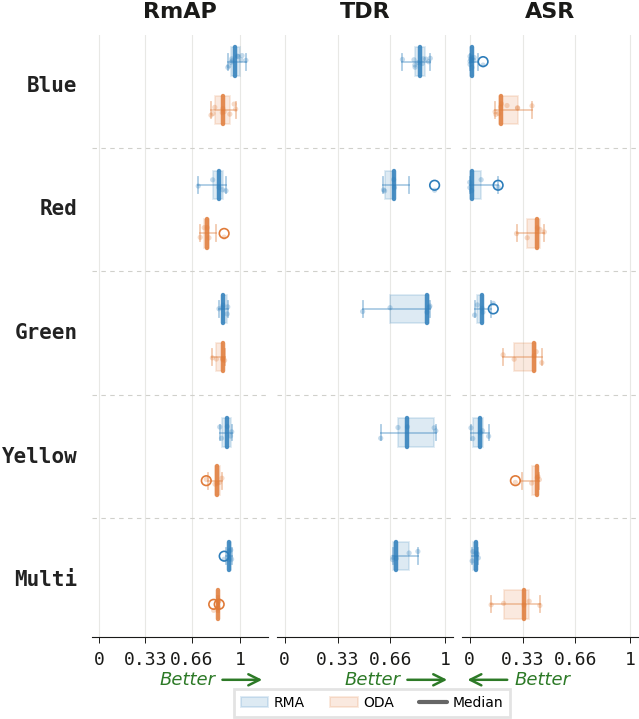}
        \caption{SBM-FWS}
        \label{fig:ptsd-rma}
    \end{subfigure}
    \hfill
    \begin{subfigure}[t]{0.45\textwidth}
        \centering
        \includegraphics[width=\linewidth]{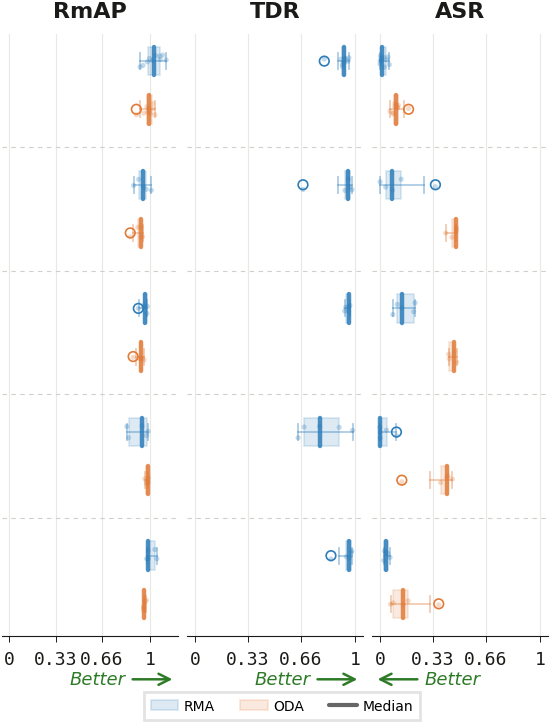}
        \caption{CLM-FWS}
        \label{fig:ptsd_oda}
    \end{subfigure}
    \centering
    \begin{subfigure}[b]{0.05\textwidth}
        \centering
        \includegraphics[width=\linewidth]{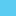}
    \end{subfigure}
    \begin{subfigure}[b]{0.05\textwidth}
        \centering
        \includegraphics[width=\linewidth]{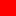}
    \end{subfigure}
    \begin{subfigure}[b]{0.05\textwidth}
        \centering
        \includegraphics[width=\linewidth]{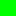}
    \end{subfigure}
    \begin{subfigure}[b]{0.05\textwidth}
        \centering
        \includegraphics[width=\linewidth]{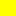}
    \end{subfigure}
    \begin{subfigure}[b]{0.05\textwidth}
        \centering
        \includegraphics[width=\linewidth]{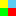}
    \end{subfigure}
\caption{CLM-FWS and SBM-FWS performance when different triggers (y-axis) are used.}
\label{fig:trigger-types}
\end{figure}

Figure~\ref{fig:trigger-types} illustrates the performance of CLM-FWS and SBM-FWS across five distinct trigger types. While our primary evaluation in Section~\ref{sec:results} utilized a blue trigger to align with PTSD dataset validation requirements, we trained DINO and FCOS models on MTSD data poisoned with various triggers to ensure our defenses are robust to trigger variations.

Overall, both defense methods demonstrate stable performance regardless of the trigger used. While RMA ASR remains largely consistent across triggers for both methods, ODA ASR exhibits more fluctuation; specifically, CLM-FWS shows greater median variance across the settings than SBM-FWS. Importantly, ODA ASR remains below 0.5 across all tested cases. Regarding TDR, SBM-FWS displays higher median variance than CLM-FWS under the RMA settings. Finally, clean performance (RmAP) for both methods remains highly stable across all trigger types.

\subsection{Pruning Threshold}
\begin{figure}
    \centering
    \includegraphics[width=\linewidth]{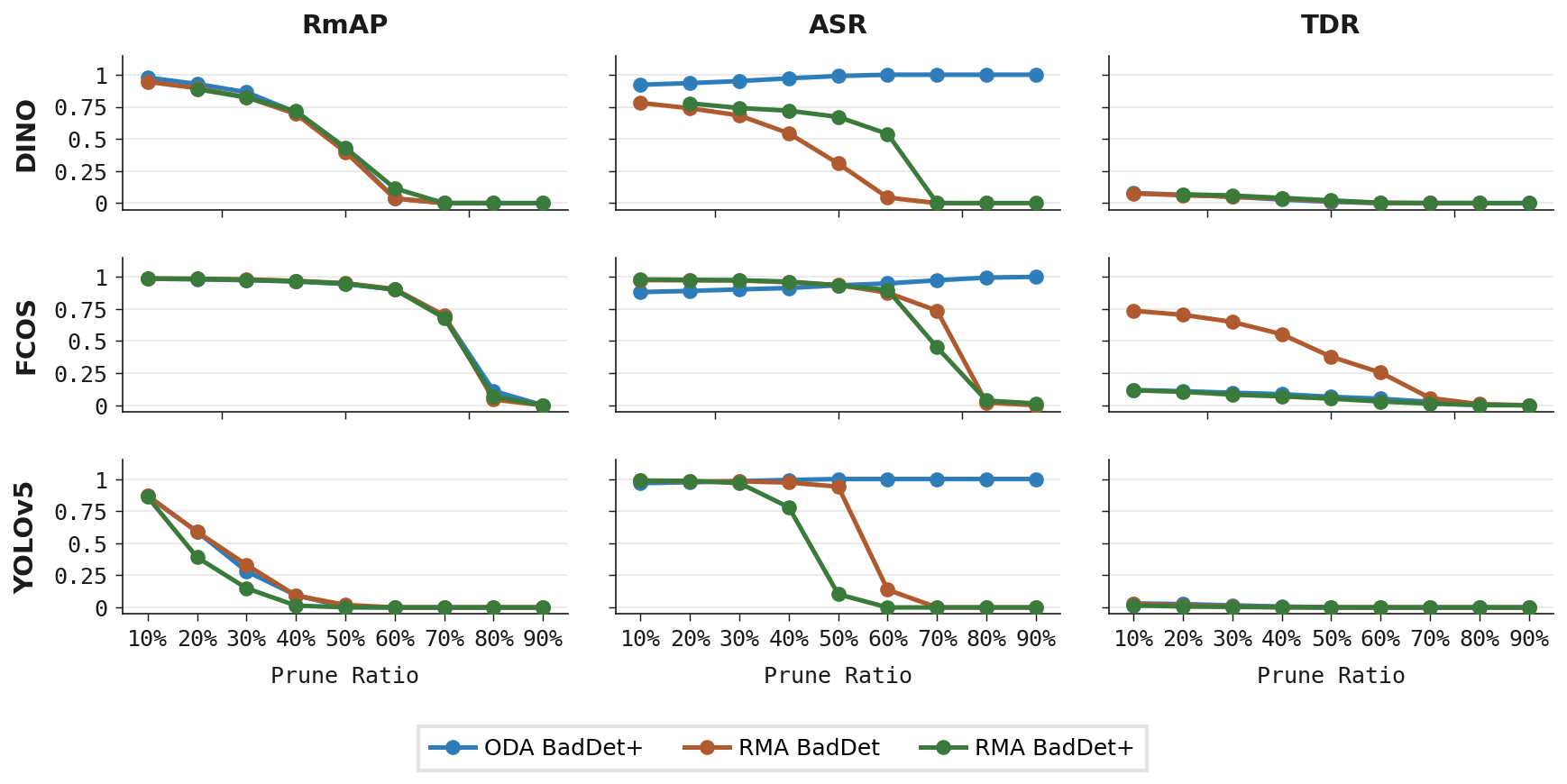}
    \caption{FP performance when pruning threshold is increased from 10 to 90\%.}
    \label{fig:prune}
\end{figure}

In Fig.~\ref{fig:prune}, we evaluate the impact of the pruning threshold ($\tau_{p}$) on Fine-Pruning (FP) performance. Across all evaluated architectures, $\tau_{p}$ fails to provide a viable trade-off between reducing ASR and maintaining clean performance (RmAP). While ASR does fall below 50\% for DINO, FCOS, and YOLOv5 at $\tau_{p}$ > 60\%, 70\%, and 50\%, respectively, this reduction is accompanied by severe RmAP degradation. Moreover, TDR does not improve at these thresholds. This indicates that the drop in ASR is a byproduct of overall detector collapse rather than effective backdoor mitigation. Furthermore, pruning is entirely ineffective against ODA-style attacks, where ASR strictly increases as $\tau_{p}$ rises.

\subsection{Additional Results}

\begin{table}[h]
  \centering
    \caption{Median performance for defense on Pascal VOC dataset using 5\% clean data.}
  \label{tab:defense_boxplot_medians_overal_voc}
  \begin{tabular}{llccc}
    \toprule
    Method & Loss & RmAP $\uparrow$ & TDR $\uparrow$ & ASR $\downarrow$ \\
    \midrule
    \multirow{2}{*}{SBM-FWS} & w/o $\mathcal{L}_{\mathrm{DEF}}$ & 0.969 & 0.711 & 0.490 \\
     & w $\mathcal{L}_{\mathrm{DEF}}$ & 0.900 & 0.676 & 0.163 \\
    \addlinespace[2pt]
    \multirow{2}{*}{SBM-RS} & w/o $\mathcal{L}_{\mathrm{DEF}}$ & 0.981 & 0.731 & 0.548 \\
     & w $\mathcal{L}_{\mathrm{DEF}}$ & 0.956 & 0.712 & 0.207 \\
    \addlinespace[2pt]
    \multirow{2}{*}{CLM-FWS} & w/o $\mathcal{L}_{\mathrm{DEF}}$ & 0.993 & 0.600 & 0.356 \\
     & w $\mathcal{L}_{\mathrm{DEF}}$ & 0.953 & 0.565 & 0.254 \\
    \addlinespace[2pt]
    \multirow{2}{*}{CLM-RS} & w/o $\mathcal{L}_{\mathrm{DEF}}$ & 0.994 & 0.653 & 0.380 \\
     & w $\mathcal{L}_{\mathrm{DEF}}$ & 0.982 & 0.630 & 0.239 \\
    \midrule
    FT & -- & 0.981 & 0.098 & 0.945 \\
    FP & -- & 0.721 & 0.031 & 0.967 \\
    FP+FT & -- & 0.972 & 0.092 & 0.947 \\
    FT-SAM & -- & 0.995 & 0.156 & 0.948 \\
    \bottomrule
  \end{tabular}
\end{table}

\begin{table}[h]
  \centering
  \caption{Median performance for defense on MTSD dataset using 5\% clean data.}
  \label{tab:defense_boxplot_medians_overall_mtsd}
  \begin{tabular}{llccc}
    \toprule
    Method & Loss & RmAP $\uparrow$ & TDR $\uparrow$ & ASR $\downarrow$ \\
    \midrule
    \multirow{2}{*}{SBM-FWS} & w/o $\mathcal{L}_{\mathrm{DEF}}$ & 0.980 & 0.892 & 0.183 \\
     & w $\mathcal{L}_{\mathrm{DEF}}$ & 0.905 & 0.861 & 0.037 \\
    \addlinespace[2pt]
    \multirow{2}{*}{SBM-RS} & w/o $\mathcal{L}_{\mathrm{DEF}}$ & 0.982 & 0.898 & 0.221 \\
     & w $\mathcal{L}_{\mathrm{DEF}}$ & 0.951 & 0.891 & 0.157 \\
    \addlinespace[2pt]
    \multirow{2}{*}{CLM-FWS} & w/o $\mathcal{L}_{\mathrm{DEF}}$ & 0.992 & 0.922 & 0.088 \\
     & w $\mathcal{L}_{\mathrm{DEF}}$ & 0.950 & 0.912 & 0.059 \\
    \addlinespace[2pt]
    \multirow{2}{*}{CLM-RS} & w/o $\mathcal{L}_{\mathrm{DEF}}$ & 1.009 & 0.917 & 0.081 \\
     & w $\mathcal{L}_{\mathrm{DEF}}$ & 0.994 & 0.916 & 0.072 \\
    \midrule
    FT & -- & 1.053 & 0.359 & 0.722 \\
    FP & -- & 1.018 & 0.132 & 0.870 \\
    FP+FT & -- & 1.066 & 0.348 & 0.747 \\
    FT-SAM & -- & 0.931 & 0.427 & 0.653 \\
    \bottomrule
  \end{tabular}
\end{table}

\begin{table}[h]
  \centering
  \caption{Median performance for defense on Pascal VOC dataset using 5\% clean data for each attack.}
  \label{tab:defense_attack_boxplot_medians_voc}
  \begin{tabular}{llccccc}
    \toprule
    & & \multicolumn{3}{c}{RMA} & \multicolumn{2}{c}{ODA} \\
    \cmidrule(lr){3-5} \cmidrule(lr){6-7}
    Method & Loss & RmAP $\uparrow$ & TDR $\uparrow$ & ASR $\downarrow$ & RmAP $\uparrow$ & ASR $\downarrow$ \\
    \midrule
    \multirow{2}{*}{SBM-FWS} & w/o $\mathcal{L}_{\mathrm{DEF}}$ & 0.968 & 0.711 & 0.649 & 0.971 & 0.364 \\
     & w $\mathcal{L}_{\mathrm{DEF}}$ & 0.897 & 0.676 & 0.126 & 0.903 & 0.379 \\
    \addlinespace[2pt]
    \multirow{2}{*}{SBM-RS} & w/o $\mathcal{L}_{\mathrm{DEF}}$ & 0.979 & 0.731 & 0.728 & 0.983 & 0.379 \\
     & w $\mathcal{L}_{\mathrm{DEF}}$ & 0.947 & 0.712 & 0.185 & 0.966 & 0.404 \\
    \addlinespace[2pt]
    \multirow{2}{*}{CLM-FWS} & w/o $\mathcal{L}_{\mathrm{DEF}}$ & 0.998 & 0.600 & 0.319 & 0.991 & 0.367 \\
     & w $\mathcal{L}_{\mathrm{DEF}}$ & 0.944 & 0.565 & 0.174 & 0.957 & 0.385 \\
    \addlinespace[2pt]
    \multirow{2}{*}{CLM-RS} & w/o $\mathcal{L}_{\mathrm{DEF}}$ & 0.994 & 0.653 & 0.355 & 0.997 & 0.387 \\
     & w $\mathcal{L}_{\mathrm{DEF}}$ & 0.982 & 0.630 & 0.225 & 0.987 & 0.378 \\
    \bottomrule
  \end{tabular}
\end{table}

\begin{table}[h]
  \centering
  \caption{Median performance for defense on MTSD dataset using 5\% clean data for each attack.}
  \label{tab:defense_attack_boxplot_medians_mtsd}
  \begin{tabular}{llccccc}
    \toprule
    & & \multicolumn{3}{c}{RMA} & \multicolumn{2}{c}{ODA} \\
    \cmidrule(lr){3-5} \cmidrule(lr){6-7}
    Method & Loss & RmAP $\uparrow$ & TDR $\uparrow$ & ASR $\downarrow$ & RmAP $\uparrow$ & ASR $\downarrow$ \\
    \midrule
    \multirow{2}{*}{SBM-FWS} & w/o $\mathcal{L}_{\mathrm{DEF}}$ & 0.967 & 0.892 & 0.239 & 0.992 & 0.138 \\
     & w $\mathcal{L}_{\mathrm{DEF}}$ & 0.894 & 0.861 & 0.014 & 0.931 & 0.251 \\
    \addlinespace[2pt]
    \multirow{2}{*}{SBM-RS} & w/o $\mathcal{L}_{\mathrm{DEF}}$ & 0.975 & 0.898 & 0.331 & 0.992 & 0.136 \\
     & w $\mathcal{L}_{\mathrm{DEF}}$ & 0.948 & 0.891 & 0.167 & 0.968 & 0.155 \\
    \addlinespace[2pt]
    \multirow{2}{*}{CLM-FWS} & w/o $\mathcal{L}_{\mathrm{DEF}}$ & 0.986 & 0.922 & 0.032 & 1.002 & 0.102 \\
     & w $\mathcal{L}_{\mathrm{DEF}}$ & 0.940 & 0.912 & 0.009 & 0.979 & 0.112 \\
    \addlinespace[2pt]
    \multirow{2}{*}{CLM-RS} & w/o $\mathcal{L}_{\mathrm{DEF}}$ & 1.009 & 0.917 & 0.057 & 1.015 & 0.091 \\
     & w $\mathcal{L}_{\mathrm{DEF}}$ & 0.989 & 0.916 & 0.020 & 1.001 & 0.100 \\
    \bottomrule
  \end{tabular}
\end{table}

\begin{table}[h]
  \centering
  \caption{Median performance for defense on Pascal VOC dataset using 5\%, 2.5\% and 1.25\% clean data}
  \label{tab:num_samples_boxplot_medians_voc}
  \begin{tabular}{lllccc}
    \toprule
    Method & Samples & Loss & RmAP $\uparrow$ & TDR $\uparrow$ & ASR $\downarrow$ \\
    \midrule
    \multirow{6}{*}{SBM} & \multirow{2}{*}{5\%} & w/o $\mathcal{L}_{\mathrm{DEF}}$ & 0.969 & 0.711 & 0.490 \\
     &  & w $\mathcal{L}_{\mathrm{DEF}}$ & 0.900 & 0.676 & 0.163 \\
    \cmidrule(l){2-6}
     & \multirow{2}{*}{2.5\%} & w/o $\mathcal{L}_{\mathrm{DEF}}$ & 0.970 & 0.695 & 0.576 \\
     &  & w $\mathcal{L}_{\mathrm{DEF}}$ & 0.880 & 0.656 & 0.239 \\
    \cmidrule(l){2-6}
     & \multirow{2}{*}{1.25\%} & w/o $\mathcal{L}_{\mathrm{DEF}}$ & 0.967 & 0.598 & 0.611 \\
     &  & w $\mathcal{L}_{\mathrm{DEF}}$ & 0.862 & 0.543 & 0.269 \\
    \midrule
    \multirow{6}{*}{CLM} & \multirow{2}{*}{5\%} & w/o $\mathcal{L}_{\mathrm{DEF}}$ & 0.993 & 0.600 & 0.356 \\
     &  & w $\mathcal{L}_{\mathrm{DEF}}$ & 0.953 & 0.565 & 0.254 \\
    \cmidrule(l){2-6}
     & \multirow{2}{*}{2.5\%} & w/o $\mathcal{L}_{\mathrm{DEF}}$ & 0.995 & 0.662 & 0.596 \\
     &  & w $\mathcal{L}_{\mathrm{DEF}}$ & 0.955 & 0.609 & 0.283 \\
    \cmidrule(l){2-6}
     & \multirow{2}{*}{1.25\%} & w/o $\mathcal{L}_{\mathrm{DEF}}$ & 0.984 & 0.591 & 0.822 \\
     &  & w $\mathcal{L}_{\mathrm{DEF}}$ & 0.935 & 0.508 & 0.376 \\
    \bottomrule
  \end{tabular}
\end{table}

\begin{table}[h]
  \centering
  \caption{Median performance for defense on MTSD dataset using 5\%, 2.5\% and 1.25\% clean data}
  \label{tab:num_samples_boxplot_medians_mtsd}
  \begin{tabular}{lllccc}
    \toprule
    Method & Samples & Loss & RmAP $\uparrow$ & TDR $\uparrow$ & ASR $\downarrow$ \\
    \midrule
    \multirow{6}{*}{SBM} & \multirow{2}{*}{5\%} & w/o $\mathcal{L}_{\mathrm{DEF}}$ & 0.980 & 0.892 & 0.183 \\
     &  & w $\mathcal{L}_{\mathrm{DEF}}$ & 0.905 & 0.861 & 0.037 \\
    \cmidrule(l){2-6}
     & \multirow{2}{*}{2.5\%} & w/o $\mathcal{L}_{\mathrm{DEF}}$ & 0.965 & 0.960 & 0.132 \\
     &  & w $\mathcal{L}_{\mathrm{DEF}}$ & 0.876 & 0.931 & 0.151 \\
    \cmidrule(l){2-6}
     & \multirow{2}{*}{1.25\%} & w/o $\mathcal{L}_{\mathrm{DEF}}$ & 0.969 & 0.834 & 0.312 \\
     &  & w $\mathcal{L}_{\mathrm{DEF}}$ & 0.918 & 0.869 & 0.108 \\
    \midrule
    \multirow{6}{*}{CLM} & \multirow{2}{*}{5\%} & w/o $\mathcal{L}_{\mathrm{DEF}}$ & 0.992 & 0.922 & 0.088 \\
     &  & w $\mathcal{L}_{\mathrm{DEF}}$ & 0.950 & 0.912 & 0.059 \\
    \cmidrule(l){2-6}
     & \multirow{2}{*}{2.5\%} & w/o $\mathcal{L}_{\mathrm{DEF}}$ & 0.939 & 0.893 & 0.101 \\
     &  & w $\mathcal{L}_{\mathrm{DEF}}$ & 0.821 & 0.893 & 0.089 \\
    \cmidrule(l){2-6}
     & \multirow{2}{*}{1.25\%} & w/o $\mathcal{L}_{\mathrm{DEF}}$ & 0.935 & 0.670 & 0.472 \\
     &  & w $\mathcal{L}_{\mathrm{DEF}}$ & 0.847 & 0.660 & 0.433 \\
    \bottomrule
  \end{tabular}
\end{table}

\newpage
\clearpage
\section*{NeurIPS Paper Checklist}

\begin{enumerate}

\item {\bf Claims}
    \item[] Question: Do the main claims made in the abstract and introduction accurately reflect the paper's contributions and scope?
    \item[] Answer: \answerYes{} % Replace by \answerYes{}, \answerNo{}, or \answerNA{}.
    \item[] Justification: The abstract and introduction state the two contributions, namely the CLM/SBM adversarial generation strategies and the dual-objective detection-aware defense loss, and the experimental scope (CNN- and Transformer-based detectors across multiple datasets and attack types) matches the empirical evaluation reported in Section~\ref{sec:results}.

\item {\bf Limitations}
    \item[] Question: Does the paper discuss the limitations of the work performed by the authors?
    \item[] Answer: \answerYes{} % Replace by \answerYes{}, \answerNo{}, or \answerNA{}.
    \item[] Justification: Limitations are discussed in Section~\ref{sec:results} and in the Conclusion, noting that fully restoring correct detections for trigger-bearing objects remains challenging (TDR@50 retains a long lower tail) and that the framework does not directly address object-generation (background-region) attacks. The theoretical analysis in Appendix~\ref{app:theory} also explicitly notes it is a mechanism-level justification rather than a global robustness guarantee.

\item {\bf Theory assumptions and proofs}
    \item[] Question: For each theoretical result, does the paper provide the full set of assumptions and a complete (and correct) proof?
    \item[] Answer: \answerYes{} % Replace by \answerYes{}, \answerNo{}, or \answerNA{}.
    \item[] Justification: The theoretical results (Propositions A1--A7) are stated with their assumptions and accompanied by full proofs in Appendix~\ref{app:theory}, which is cross-referenced throughout the design discussion in Sections~\ref{sec:advgen}--\ref{sec:overall_opt}.

    \item {\bf Experimental result reproducibility}
    \item[] Question: Does the paper fully disclose all the information needed to reproduce the main experimental results of the paper to the extent that it affects the main claims and/or conclusions of the paper (regardless of whether the code and data are provided or not)?
    \item[] Answer: \answerYes{} % Replace by \answerYes{}, \answerNo{}, or \answerNA{}.
    \item[] Justification: Section~\ref{sec:results} and Appendix~\ref{sec:appendix-selection-hyper} (with Table~\ref{tab:hyperparameters}) report all per-architecture hyperparameters, optimizer settings, the number of epochs, batch size, adversarial steps, step size, data splits (5\%, 2.5\%, 1.75\% with five random splits), and trigger configurations needed to reproduce the main results, alongside the publicly available datasets and architectures used.

\item {\bf Open access to data and code}
    \item[] Question: Does the paper provide open access to the data and code, with sufficient instructions to faithfully reproduce the main experimental results, as described in supplemental material?
    \item[] Answer: \answerYes{} % Replace by \answerYes{}, \answerNo{}, or \answerNA{}.
    \item[] Justification: As stated in Section~\ref{sec:results}, the benchmarking framework code is included with the submission and will be released publicly on GitHub upon acceptance. All datasets used (Pascal VOC, MTSD, COCO, PTSD) are publicly available from their original providers.

\item {\bf Experimental setting/details}
    \item[] Question: Does the paper specify all the training and test details (e.g., data splits, hyperparameters, how they were chosen, type of optimizer) necessary to understand the results?
    \item[] Answer: \answerYes{} % Replace by \answerYes{}, \answerNo{}, or \answerNA{}.
    \item[] Justification: The experimental setup (datasets, attack settings, baselines, metrics, and clean-data budgets) is described in Section~\ref{sec:results}, and full training details (learning rate, epochs, batch size, adversarial steps and step size, $\lambda$, $\beta$, $\epsilon$, IoU threshold, matching strategy) are provided in Table~\ref{tab:hyperparameters} of Appendix~\ref{sec:appendix-selection-hyper}, with sensitivity analyses justifying how each hyperparameter was chosen.

\item {\bf Experiment statistical significance}
    \item[] Question: Does the paper report error bars suitably and correctly defined or other appropriate information about the statistical significance of the experiments?
    \item[] Answer: \answerYes{} % Replace by \answerYes{}, \answerNo{}, or \answerNA{}.
    \item[] Justification: All main results (Figures~\ref{fig:overall_results}, \ref{fig:attack_type}, \ref{fig:num_samples}, and the appendix figures) are reported as box plots that show median, interquartile range, and the overall distribution computed across five random data splits, capturing variability due to the choice of clean-data subset.

\item {\bf Experiments compute resources}
    \item[] Question: For each experiment, does the paper provide sufficient information on the computer resources (type of compute workers, memory, time of execution) needed to reproduce the experiments?
    \item[] Answer: \answerYes{} % Replace by \answerYes{}, \answerNo{}, or \answerNA{}.
    \item[] Justification: Section~\ref{sec:results} and the timing analysis in Appendix (Table~\ref{tab:timing}) report that experiments were conducted on a single H100 GPU and provide a per-batch timing breakdown across FCOS, DINO, and YOLOv5 for each defence variant.
    
\item {\bf Code of ethics}
    \item[] Question: Does the research conducted in the paper conform, in every respect, with the NeurIPS Code of Ethics \url{https://neurips.cc/public/EthicsGuidelines}?
    \item[] Answer: \answerYes{} % Replace by \answerYes{}, \answerNo{}, or \answerNA{}.
    \item[] Justification: The research uses publicly available benchmark datasets, does not involve human subjects, and is aimed at improving the security of object detectors against backdoor attacks; we have reviewed the NeurIPS Code of Ethics and believe our work conforms to it.

\item {\bf Broader impacts}
    \item[] Question: Does the paper discuss both potential positive societal impacts and negative societal impacts of the work performed?
    \item[] Answer: \answerYes{} % Replace by \answerYes{}, \answerNo{}, or \answerNA{}.
    \item[] Justification: The Introduction motivates the work in terms of safety-critical applications such as autonomous driving and surveillance, where backdoor mitigation in object detectors yields direct positive impact; because the contribution is a defensive mitigation method (not a new attack or generative capability), the path to direct negative societal impact is limited.
    
\item {\bf Safeguards}
    \item[] Question: Does the paper describe safeguards that have been put in place for responsible release of data or models that have a high risk for misuse (e.g., pre-trained language models, image generators, or scraped datasets)?
    \item[] Answer: \answerNA{} % Replace by \answerYes{}, \answerNo{}, or \answerNA{}.
    \item[] Justification: The paper does not release any pre-trained generative models, scraped data, or other assets that pose a high risk of misuse; the released artifact is a defensive backdoor-mitigation framework built on existing public detectors and benchmark datasets.

\item {\bf Licenses for existing assets}
    \item[] Question: Are the creators or original owners of assets (e.g., code, data, models), used in the paper, properly credited and are the license and terms of use explicitly mentioned and properly respected?
    \item[] Answer: \answerYes{} % Replace by \answerYes{}, \answerNo{}, or \answerNA{}.
    \item[] Justification: All datasets (Pascal VOC, MTSD, COCO, PTSD) and detector architectures (DINO, FCOS, YOLOv5) used in the paper are cited to their original publications/sources in Section~\ref{sec:results}, and they are used in accordance with their respective public terms of use.

\item {\bf New assets}
    \item[] Question: Are new assets introduced in the paper well documented and is the documentation provided alongside the assets?
    \item[] Answer: \answerYes{} % Replace by \answerYes{}, \answerNo{}, or \answerNA{}.
    \item[] Justification: The new asset is the benchmarking and mitigation code, which is included with the submission and will be released on GitHub upon acceptance; method, hyperparameters, and design choices are documented in Section~\ref{sec:results} and Appendices~\ref{sec:appendix-selection-hyper} and~\ref{sec:appendix-additional-ablations}.

\item {\bf Crowdsourcing and research with human subjects}
    \item[] Question: For crowdsourcing experiments and research with human subjects, does the paper include the full text of instructions given to participants and screenshots, if applicable, as well as details about compensation (if any)? 
    \item[] Answer: \answerNA{} % Replace by \answerYes{}, \answerNo{}, or \answerNA{}.
    \item[] Justification: The paper does not involve any crowdsourcing or research with human subjects; all experiments are conducted on existing public benchmark datasets.

\item {\bf Institutional review board (IRB) approvals or equivalent for research with human subjects}
    \item[] Question: Does the paper describe potential risks incurred by study participants, whether such risks were disclosed to the subjects, and whether Institutional Review Board (IRB) approvals (or an equivalent approval/review based on the requirements of your country or institution) were obtained?
    \item[] Answer: \answerNA{} % Replace by \answerYes{}, \answerNo{}, or \answerNA{}.
    \item[] Justification: The paper does not involve crowdsourcing or research with human subjects, so IRB approval is not applicable.

\item {\bf Declaration of LLM usage}
    \item[] Question: Does the paper describe the usage of LLMs if it is an important, original, or non-standard component of the core methods in this research? Note that if the LLM is used only for writing, editing, or formatting purposes and does \emph{not} impact the core methodology, scientific rigor, or originality of the research, declaration is not required.
    %this research? 
    \item[] Answer: \answerNA{} % Replace by \answerYes{}, \answerNo{}, or \answerNA{}.
    \item[] Justification: LLMs are not used as a component of the core methodology; the work concerns adversarial fine-tuning of object detectors (FCOS, DINO, YOLOv5) and does not rely on large language models in any important, original, or non-standard way.

\end{enumerate}

\end{document}